\documentclass[10pt,journal,compsoc]{IEEEtran}
\ifCLASSOPTIONcompsoc
  \usepackage[nocompress]{cite}
\else
  \usepackage{cite}
\fi
\ifCLASSINFOpdf
\else
\fi

\usepackage{amsmath,amsfonts,amssymb}
\usepackage{mathtools}
\usepackage{amsthm}
\usepackage{subcaption}
\usepackage{alltt, xspace, times, epsfig, url}
\usepackage{gensymb,xfrac,array,booktabs,bm,xcolor}
\usepackage[linesnumbered,lined,ruled]{algorithm2e}
\usepackage{multirow}
\usepackage{multicol}
\usepackage{pifont}
\usepackage{algorithmic}

\begin{document}
%\title{XEnsemble: Robust Deep Learning Ensemble against Adversarial Examples and Out-of-Distribution Attacks}
\title{\huge Robust Deep Learning Ensemble against Deception}

\author{Wenqi~Wei,~\IEEEmembership{Student member,~IEEE}, and~Ling~Liu,~\IEEEmembership{Fellow,~IEEE}
% <-this % stops a space
% \IEEEcompsocitemizethanks{\IEEEcompsocthanksitem The authors acknowledge the partial support from the National Science Foundation under Grants SaTC 1564097, NSF 1547102, and an IBM Faculty Award.}

\IEEEcompsocitemizethanks{\IEEEcompsocthanksitem Wenqi Wei and Ling Liu are with School of Computer Science, Georgia Institute of Technology, Atlanta,
GA, 30332.\protect\\
E-mail: wenqiwei@gatech.edu, ling.liu@cc.gatech.edu}% <-this % stops an unwanted space
%\hfil\break # for a new line
\thanks{Manuscript received xxxx xx, xxxx; revised xxxxx xx, xxxx.}
}

%Ka-Ho~Chow, Yanzhao~Wu,
%K. Chow, Y. Wu,
%,khchow,yanzhaowu


\markboth{IEEE Transactions on Dependable and Secure Computing,~Vol.~xx, No.~x, xxxx~20xx}%
{Shell \MakeLowercase{\textit{et al.}}: Bare Demo of IEEEtran.cls for Computer Society Journals}



\IEEEtitleabstractindextext{%
\begin{abstract}
Deep neural network (DNN) models are known to be vulnerable to maliciously crafted adversarial examples and to out-of-distribution inputs drawn sufficiently far away from the training data. How to protect a machine learning model against deception of both types of destructive inputs remains an open challenge. This paper presents XEnsemble,  a diversity ensemble verification methodology for enhancing the adversarial robustness of DNN models against deception caused by either adversarial examples or out-of-distribution inputs.
XEnsemble by design has three unique capabilities. First, XEnsemble builds diverse input denoising verifiers by leveraging different data cleaning techniques. Second, XEnsemble develops a disagreement-diversity ensemble learning methodology for guarding the output of the prediction model against deception. Third, XEnsemble provides a suite of algorithms to combine input verification and output verification to protect the DNN prediction models from both adversarial examples and out of distribution inputs. Evaluated using eleven popular adversarial attacks and two representative out-of-distribution datasets, we show that XEnsemble achieves a high defense success rate against adversarial examples and a high detection success rate against out-of-distribution data inputs, and outperforms existing representative defense methods with respect to robustness and defensibility.
%Moreover, we present the limitation of the proposed defense under adaptive knowledgeable attackers and propose the possible mitigation strategy.
\end{abstract}

% \begin{IEEEkeywords}
% Computer Society, IEEE, IEEEtran, journal, \LaTeX, paper, template.
% \end{IEEEkeywords}
}

\maketitle

\IEEEdisplaynontitleabstractindextext

% \fi
\IEEEpeerreviewmaketitle

\IEEEraisesectionheading{\section{Introduction}\label{sec:introduction}}

\IEEEPARstart{D}{eep} learning has achieved unparalleled success across a variety of machine learning tasks, e.g., image classification and speech recognition.
However, deep neural network (DNN) models are also known to be vulnerable to deceptive inputs that are either adversarial examples or out-of-distribution examples. Adversarial examples~\cite{goodfellow6572explaining} are malicious inputs crafted covertly over benign examples to fool deep learning models to misclassify without being detected by the human, and out-of-distribution examples~\cite{amodei2016concrete} are query inputs whose distribution differs from the training distribution. Both problems pose potential threats to many mission-critical systems and applications that use deep learning as a primary decision-making component, such as real-time object recognition, self-driving cars and voice command recognition.

%sharif2016accessorize, ~\cite{krizhevsky2012imagenet} ~\cite{hinton2012deep}



The robustness of a DNN model depends on its capability to survive from the deception vulnerabilities due to both maliciously crafted adversarial inputs and out-of-distribution inputs. Both types of deceptive inputs are attributed to the same problem: the input to the prediction model has been altered maliciously.
Existing defense methods against adversarial examples do not generalize over different attack algorithms~\cite{goodfellow2018defense}. Different defense methods tend to have different robustness under either different attack algorithms or different settings of attack parameters of the same attack algorithm~\cite{wei2020adversarial}. Meanwhile, existing out-of-distribution defenses either have a limited defense success rate~\cite{lee2018simple} on the adversarial example or fail miserably~\cite{liang2017enhancing}. In this paper, we make two arguments. First, a robust defense of a privately trained DNN model against deception should be capable of shielding the well-trained prediction model from intentional manipulation with either maliciously patched inputs or out-of-distribution inputs. Second, a robust defense solution should be independent of attacks and should generalize well across different attacks, different datasets, and different DNN algorithms.

\begin{figure*}[ht]
\begin{minipage}{0.99\linewidth}
 \centerline{\includegraphics[scale=.76]{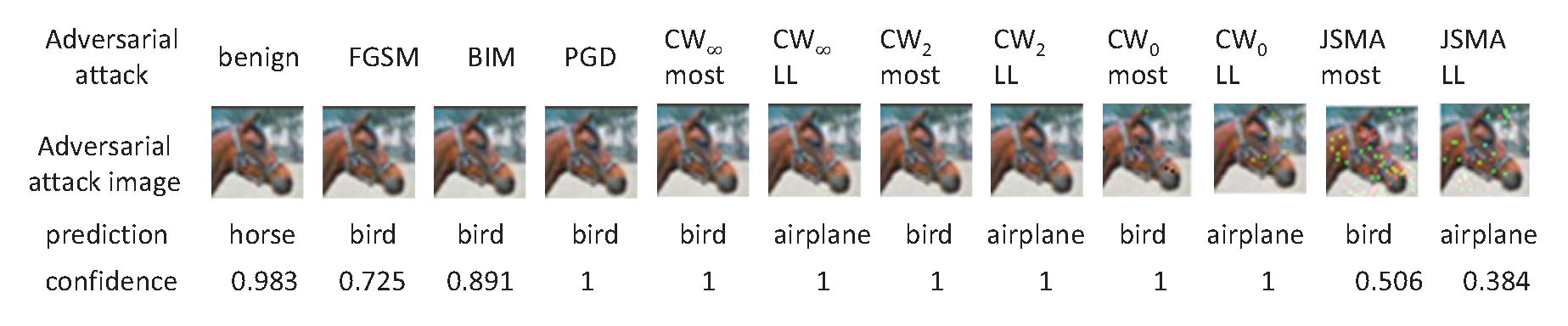}}
 \vspace{-0.2cm}
 \subcaption{\small Adversarial attacks}
 \label{figure:adv_attack}
\end{minipage}
\begin{minipage}{0.99\linewidth}
 \centerline{\includegraphics[scale=.76]{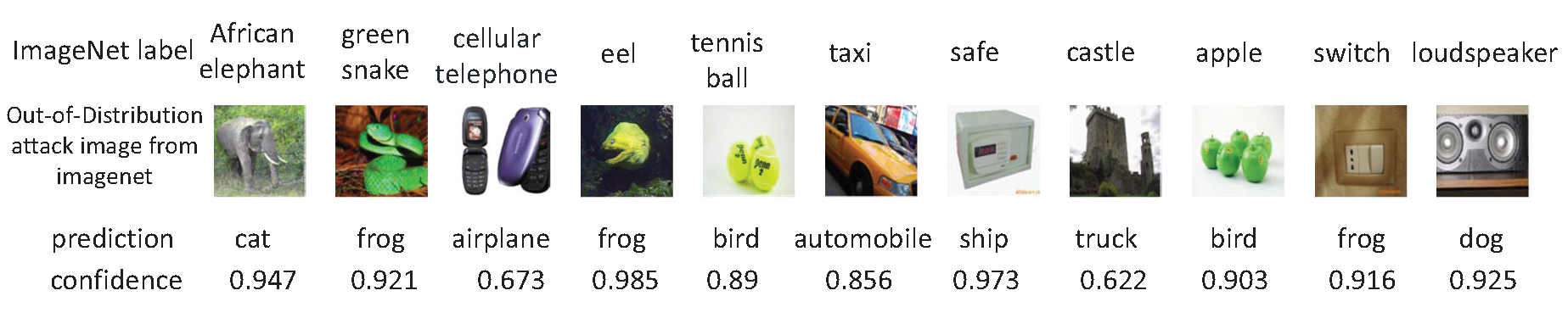}}
 \vspace{-0.2cm}
  \subcaption{\small Out-of-distribution attack}
   \label{figure:ood_attack}
\end{minipage}
\caption{\small Visualization of adversarial attacks and the OOD attack using TinyImageNet to a CIFAR-10 DenseNet classifier.}
\label{figure:attack_visualization}
\vspace{-0.4cm}
\end{figure*}

Bearing these problems in mind, we develop a diversity ensemble verification methodology, called XEnsemble, to maximize the adversarial robustness of DNN models against deception. The design of XEnsemble leverages two unique properties of deceptive inputs: (1) if an ensemble consists of only weak models whose prediction accuracy is lower than the victim target model, which we aim to protect, then we show that such an ensemble of weak models will be a poor ensemble choice for defending against deceptive inputs. (2) Not all ensemble teams from well-trained models can be effective against deception. We show that only those ensembles that have high fault tolerance in terms of disagreement diversity and respond to negative examples very differently are the ensemble teams of high robustness. A distinct property of high disagreement-diversity ensembles is that their member models tend to have very different gradient information. This contrasting gradient information tends to cause the same adversarial examples to have highly divergent and inconsistent behavior across their member models. Furthermore, this conflicting gradient information provides a distinctive opportunity for a disagreement diversity ensemble to integrate diverse decision boundaries from its member models to infer the ensemble decision boundary, which consequently reduces the attack transferability~\cite{papernot2016transferability} of both adversarial and out-of-distribution inputs.

In summary, XEnsemble by design is capable of auto-verifying any input to the prediction model to minimize the detrimental effects due to malicious and erroneous data inputs. Consider that adversarial examples are typically generated from their corresponding benign examples, XEnsemble should be capable of auto-repairing the adverse effects caused by adversarial examples generated by different attack methods, aiming to maximize the repairing rate for adversarial examples and to achieve a high prevention success rate (PSR). For those adversarial examples that escape from the auto-repairing process, we seek to
auto-detect them and filter them out.
At the same time, since the out-of-distribution inputs are drawn far away from the class distribution of a trained prediction model and are outliers to the prediction tasks, the defense should maximize the detection success rate (TSR) for identifying as many out-of-distribution examples as possible and prune them out.

This paper makes three novel contributions.
First, the proposed XEnsemble approach is attack-independent. It verifies the input to the prediction model by using multiple diverse input data cleansing techniques to denoise primarily the adversarial inputs. Then it verifies the output of the attack target model by creating multiple failure-independent model verifiers that have high disagreement-diversity.
Second, the proposed XEnsemble approach can effectively increase the adversarial robustness of the attack target DNN model by maximizing the number of adversarial inputs that can be auto-repaired and maximizing the number of out-of-distribution inputs that can be detected.
Third but not the least, our XEnsemble algorithms for auto-verification, auto-repairing, and auto-detection can generalize well across attack methods since XEnsemble algorithms do not require re-training the DNN model under protection using existing attack algorithms, and can be easily implemented on modern neural network architectures and used as a plug-in by the state of the art deep learning software frameworks. Evaluated using eleven popular adversarial attacks and two representative out-of-distribution datasets, we show that XEnsemble approach can effectively strengthen the robustness of a DNN model under protection in terms of high attack prevention success rate (PSR) for adversarial examples and high attack detection success rate(TSR) for out-of-distribution input when comparing to existing representative defense techniques. It is also worth noting that the DNN model under the protection of XEnsemble can further improve its prediction accuracy in the presence of no attacks.

% The rest of the paper is organized as follows. Section 2 provides an overview of the deception threat and existing defense approaches. Section 3 presents the solution approach of XEnsemble defense and Section 4 provides the evaluation and comparison of the proposed defense on adversarial examples and out-of-distribution input. Section~5 talks about the limitation and optimization of the defensibility and section~6 concludes the paper.

\section{Overview and Problem Statement}

\subsection{Deceptive Input Artifacts}
{\bf Adversarial Examples\/} are the input artifacts that are created from benign inputs by adding adversarial distortions, aiming to cause the target model to misclassify with high confidence~\cite{szegedy2013intriguing}. Adversarial example generation process can be formulated as
$$min||x - x'|{|_p} \quad s.t.f(x') = {y^*},\ f(x') \ne f(x),$$
where $y^{*}$ denotes the target class label for targeted attacks or any label other than the correct class for untargeted attacks. $p$ is the distance norm, such as $L_0$, $L_2$, and $L_\infty$.
For a given adversarial input $x_{adv}$ of a benign image $x$, $L_0$ norm indicates the number of pixels of $x$ that are changed in $x_{adv}$, $L_2$ norm is the Euclidean distance between $x$ and $x_{adv}$, and $L_\infty$ norm denotes the maximum change to any pixel of benign input $x$ in $x_{adv}$. For targeted attacks, we consider two targets representing two ends of the attack spectrum: the most-likely attack class in the prediction vector ($y^T = \arg \max \nolimits_{y \neq C_x} \overrightarrow y$, most) and the least-likely attack class ($y^T = \arg \min \overrightarrow y$, LL). $C_x$ is the correctly predicted label for benign input $x$.
We use eleven attack algorithms from seven families of attack algorithms to evaluate and compare the performance of defense approaches, including FGSM~\cite{szegedy2013intriguing}, BIM~\cite{kurakin2016physical}, PGD~\cite{madry2017towards}, CW$_0$, CW$_2$, CW$_\infty$~\cite{carlini2017towards}, and JSMA~\cite{papernot2016limitations}. Figure~\ref{figure:adv_attack} provides a visualization of adversarial perturbation attack to a DenseNet classifier well-trained on CIFAR-10. Its prediction on the benign horse input is a horse with 0.983 confidence. Each adversarial example generated by one of the 11 attacks causes the target model to misclassify the horse image into either a bird or an airplane, with some attacks at high confidence of 1.

{\bf Out-of-Distribution(OOD) Inputs\/} are the artifacts of abnormal inputs drawn from a completely different distribution than the domain of the learning task. OOD inputs can be easily utilized by an adversary to launch malicious attacks because when deploying neural networks in real-world applications, there is often very little control over the test data distribution~\cite{amodei2016concrete}.
%In this case, the ten categories of images represented by the CIFAR-10 dataset. OOD inputs are practically possible when deploying neural networks in real-world applications.
%%
%%Although we can always retrain the model and add new labels, the training can be more and more computationally expensive and intractable as the number of classes grows.
%%
Figure~\ref{figure:ood_attack} shows eleven OOD examples, which are taken from ImageNet and fed into a CIFAR-10 DNN classifier. We observe that the cell phone image is misclassified as an airplane and the apple image is misclassified as a bird because the DNN classifier is trained using DenseNet on
the CIFAR-10 dataset. However, OODs are considered ill-input data to this well-trained model, even though the DenseNet CIFAR-10 classifier has 94.8\% benign test accuracy. The OODs are either completely unrecognizable~\cite{nguyen2015deep} or totally irrelevant to the scope of the given learning task~\cite{hendrycks2016baseline}. For conventional computing programs, the ill-input data is known to cause detrimental effect either during the runtime execution or to the output of a program. Thus, input and output verification mechanisms are typically employed at both compile time and runtime to ensure their correctness. In this paper, we argue that there should be no exception to the input and output of an executable prediction model. XEnsemble presents a holistic approach to integrating input and output ensemble learning based verifications for improving the robustness of deep learning against deception.

\subsection{Adversarial Attack Algorithms}

Let $x$ and $x_{adv}$ be a benign input and its adversarial example respectively, f(x; $\theta$) denote the prediction model with $\theta$ representing the neural network parameters, and f(x; $\theta$) takes an input query example $x$ and outputs $\overrightarrow {y}$ and $y_{x}$ as the prediction vector and predicted class for the benign input $x$ respectively. During the model training phase, $f(x, C_{x};\theta)$ is iteratively learned over the fixed training dataset of $\{x, C_{x}\}$, where $C_{x}$ denotes the class label given for training input $x$. The model parameters $\theta$ are updated through gradient descent optimization using back-propagation on a loss function, denoted by $J(h_{i}(x),C_{x})$ where $h_{i}(x)$ is the intermediate result at the $i$th iteration and the training completes when the convergence condition of the iterative training is met, which produces $f(x; \theta)$ as the resulting prediction model. At the prediction phase, the prediction model $f(x; \theta)$ takes any query input $x$ and produces the prediction output denoted by $\overrightarrow {y}$ and $y_{x}$.

When adversarial examples are sent to a prediction model, the goal of untargeted attack~\cite{wei2020adversarial} is to cause the prediction model to misclassify. Thus, with $x_{adv}=x+\Delta_x$, the attack objective is to perturb the benign  input $x$ with small $\Delta_x$ to increase the loss function value $J(h(x+\Delta_x), y_{x})$ such that the prediction result will be altered from the correct class label $y_{x}$ of its benign input $x$. By gradient ascent, the attacker needs to decrease the value of $x$ when the partial gradient is below 0 or to increase the value of $x$ when the partial gradient is above 0, which will result in increasing the loss value of $J(h(x+\Delta_x),y_{x})$.

For targeted attacks, the goal of the attacker is changed to minimize the value of the loss function $J(h(x+\Delta_x),y_t)$ where $y_t$ denotes the attack target class label. Thus, the attacker needs to decrease the pixel value of benign example $x$ when the partial gradient is above 0 or to increase the pixel value of $x$ when the partial gradient is below 0, which will cause to the value of the loss function $J(h(x+\Delta_x),y_{x})$ to decrease.

White-box attack v.s. Black-box attack. White-box attacks refer to adversarial examples generated by attackers who have full knowledge of the prediction model. All attacks that manipulate the gradient Information are white-box attacks since attackers use the gradient information of the attack objective to guide the learning of adversarial perturbation in terms of where (location) and how much (amount) noise should be injected to a benign input $x$. Black-box attacks refer to adversarial examples that are generated from a surrogate model that an attack generates, which can be viewed as a shadow model of the private attack target model. By the transferability of adversarial examples~\cite{papernot2017practical,papernot2016transferability}, one can launch black-box attacks using such adversarial examples. One approach to generating the surrogate model for any privately held prediction model is to utilize query probing through the model prediction API and the membership inference~\cite{truex2019demystifying}.

\begin{table}[ht]
\centering
\scalebox{0.76}{
\small{
\begin{tabular}{ccccccccc}
\hline
\multicolumn{1}{|c|}{\multirow{13}{*}{\rotatebox{90}{CIFAR-10}}}& \multicolumn{2}{c|}{\multirow{2}{*}{attack}}                                                          & \multicolumn{2}{c|}{attack effect}                     & \multicolumn{1}{c|}{confidence}                           & \multicolumn{3}{c|}{cost}                                                                           \\  \cline{4-9}
\multicolumn{1}{|c|}{} &\multicolumn{2}{c|}{}                                                            & \multicolumn{1}{c|}{ASR} & \multicolumn{1}{c|}{MR} & \multicolumn{1}{c|}{AdvConf} & \multicolumn{1}{c|}{Perturb} & \multicolumn{1}{c|}{Percept} & \multicolumn{1}{c|}{Time(s)} \\ \cline{2-9}
\multicolumn{1}{|c|}{} &\multicolumn{1}{c|}{FGSM}                   & \multicolumn{1}{c|}{\multirow{3}{*}{UA}} & \multicolumn{1}{c|}{0.85}        & \multicolumn{1}{c|}{0.85}   & \multicolumn{1}{c|}{0.8647}      & \multicolumn{1}{c|}{0.93}        & \multicolumn{1}{c|}{\textbf{47.63}}       & \multicolumn{1}{c|}{0.021}    \\ \cline{2-2} \cline{4-9}
\multicolumn{1}{|c|}{} &\multicolumn{1}{c|}{BIM}                    & \multicolumn{1}{c|}{}                    & \multicolumn{1}{c|}{0.92}        & \multicolumn{1}{c|}{0.92}   & \multicolumn{1}{c|}{0.9645}      & \multicolumn{1}{c|}{0.607}       & \multicolumn{1}{c|}{18.96}       & \multicolumn{1}{c|}{0.154}    \\ \cline{2-2} \cline{4-9}
\multicolumn{1}{|c|}{} &\multicolumn{1}{c|}{PGD}                    & \multicolumn{1}{c|}{}                    & \multicolumn{1}{c|}{0.86}        & \multicolumn{1}{c|}{0.86}   & \multicolumn{1}{c|}{0.9629}      & \multicolumn{1}{c|}{0.2}       & \multicolumn{1}{c|}{27.67}       & \multicolumn{1}{c|}{0.44}   \\ \cline{2-2}  \cline{2-9}
\multicolumn{1}{|c|}{} &\multicolumn{1}{c|}{\multirow{2}{*}{CW$_\infty$}} & \multicolumn{1}{c|}{most}                  & \multicolumn{1}{c|}{1}       & \multicolumn{1}{c|}{1}   & \multicolumn{1}{c|}{0.9889}      & \multicolumn{1}{c|}{0.571}       & \multicolumn{1}{c|}{15.98}       & \multicolumn{1}{c|}{\textbf{235.5}}    \\ \cline{3-9}
\multicolumn{1}{|c|}{} &\multicolumn{1}{c|}{}                       & \multicolumn{1}{c|}{LL}                  & \multicolumn{1}{c|}{1}       & \multicolumn{1}{c|}{1}  & \multicolumn{1}{c|}{0.9779}      & \multicolumn{1}{c|}{0.726}       & \multicolumn{1}{c|}{26.45}       & \multicolumn{1}{c|}{\textbf{243.2}}    \\ \cline{2-9}
\multicolumn{1}{|c|}{} &\multicolumn{1}{c|}{\multirow{2}{*}{CW$_2$}} & \multicolumn{1}{c|}{most}                  & \multicolumn{1}{c|}{1}       & \multicolumn{1}{c|}{1}   & \multicolumn{1}{c|}{0.9867}      & \multicolumn{1}{c|}{0.455}       & \multicolumn{1}{c|}{6.92}        & \multicolumn{1}{c|}{5.772}    \\ \cline{3-9}
\multicolumn{1}{|c|}{} &\multicolumn{1}{c|}{}                       & \multicolumn{1}{c|}{LL}                  & \multicolumn{1}{c|}{1}       & \multicolumn{1}{c|}{1}   & \multicolumn{1}{c|}{0.9732}      & \multicolumn{1}{c|}{0.598}       & \multicolumn{1}{c|}{13}          & \multicolumn{1}{c|}{7.441}    \\ \cline{2-9}
\multicolumn{1}{|c|}{}&\multicolumn{1}{c|}{\multirow{2}{*}{CW$_0$}} & \multicolumn{1}{c|}{most}                  & \multicolumn{1}{c|}{1}       & \multicolumn{1}{c|}{1}  & \multicolumn{1}{c|}{0.9904}      & \multicolumn{1}{c|}{\textbf{1.251}}       & \multicolumn{1}{c|}{8.003}       & \multicolumn{1}{c|}{\textbf{355.4}}    \\ \cline{3-9}
\multicolumn{1}{|c|}{}&\multicolumn{1}{c|}{}                       & \multicolumn{1}{c|}{LL}                  & \multicolumn{1}{c|}{1}       & \multicolumn{1}{c|}{1}  & \multicolumn{1}{c|}{0.9757}      & \multicolumn{1}{c|}{\textbf{1.587}}       & \multicolumn{1}{c|}{18.11}       & \multicolumn{1}{c|}{\textbf{356.7}}    \\ \cline{2-9}
\multicolumn{1}{|c|}{}&\multicolumn{1}{c|}{\multirow{2}{*}{JSMA}}  & \multicolumn{1}{c|}{most}                  & \multicolumn{1}{c|}{1}       & \multicolumn{1}{c|}{1}   & \multicolumn{1}{c|}{0.5366}      & \multicolumn{1}{c|}{\textbf{1.934}}       & \multicolumn{1}{c|}{27.12}       & \multicolumn{1}{c|}{4.894}    \\ \cline{3-9}
\multicolumn{1}{|c|}{}&\multicolumn{1}{c|}{}                       & \multicolumn{1}{c|}{LL}                  & \multicolumn{1}{c|}{0.99}        & \multicolumn{1}{c|}{1}  & \multicolumn{1}{c|}{0.3920}      & \multicolumn{1}{c|}{\textbf{2.338}}       & \multicolumn{1}{c|}{\textbf{53.48}}       & \multicolumn{1}{c|}{9.858}    \\ \hline
\multicolumn{1}{|c}{} &\multicolumn{1}{c}{}                       & \multicolumn{1}{c}{}           & \multicolumn{1}{c}{}           & \multicolumn{1}{c}{}           & \multicolumn{1}{c}{}          & \multicolumn{1}{c}{}           & \multicolumn{1}{c}{}           & \multicolumn{1}{c|}{}  \\ \hline
\multicolumn{1}{|c|}{\multirow{9}{*}{\rotatebox{90}{ImageNet}}} & \multicolumn{1}{c|}{FGSM}                   & \multicolumn{1}{c|}{\multirow{3}{*}{UA}} & \multicolumn{1}{c|}{0.99}        & \multicolumn{1}{c|}{0.99}    & \multicolumn{1}{c|}{0.6408}      & \multicolumn{1}{c|}{1.735}       & \multicolumn{1}{c|}{\textbf{1152}}    & \multicolumn{1}{c|}{0.019}    \\ \cline{2-2} \cline{4-9}
\multicolumn{1}{|c|}{} &\multicolumn{1}{c|}{BIM}                    & \multicolumn{1}{c|}{}                    & \multicolumn{1}{c|}{1}       & \multicolumn{1}{c|}{1}   & \multicolumn{1}{c|}{0.9971}      & \multicolumn{1}{c|}{1.186}       & \multicolumn{1}{c|}{502.5}       & \multicolumn{1}{c|}{0.185}\\  \cline{2-2} \cline{4-9}
\multicolumn{1}{|c|}{} &\multicolumn{1}{c|}{PGD}            & \multicolumn{1}{c|}{}         & \multicolumn{1}{c|}{1}                    & \multicolumn{1}{c|}{1}        & \multicolumn{1}{c|}{0.9986}   & \multicolumn{1}{c|}{1.636}      & \multicolumn{1}{c|}{669.2}       & \multicolumn{1}{c|}{0.536}         \\ \cline{2-9}
\multicolumn{1}{|c|}{} &\multicolumn{1}{c|}{\multirow{2}{*}{CW$_\infty$}} & \multicolumn{1}{c|}{most}                  & \multicolumn{1}{c|}{1}       & \multicolumn{1}{c|}{1}  & \multicolumn{1}{c|}{0.9850}      & \multicolumn{1}{c|}{0.957}       & \multicolumn{1}{c|}{217.9}       & \multicolumn{1}{c|}{74.7}     \\ \cline{3-9}
\multicolumn{1}{|c|}{} &\multicolumn{1}{c|}{}                       & \multicolumn{1}{c|}{LL}                  & \multicolumn{1}{c|}{0.95}        & \multicolumn{1}{c|}{0.96}   & \multicolumn{1}{c|}{0.8155}      & \multicolumn{1}{c|}{1.394}       & \multicolumn{1}{c|}{592.8}       & \multicolumn{1}{c|}{\textbf{237.8}}    \\ \cline{2-9}
\multicolumn{1}{|c|}{} &\multicolumn{1}{c|}{\multirow{2}{*}{CW$_2$}} & \multicolumn{1}{c|}{most}                  & \multicolumn{1}{c|}{1}       & \multicolumn{1}{c|}{1}  & \multicolumn{1}{c|}{0.9069}      & \multicolumn{1}{c|}{0.836}       & \multicolumn{1}{c|}{120.5}       & \multicolumn{1}{c|}{13.2}     \\ \cline{3-9}
\multicolumn{1}{|c|}{} &\multicolumn{1}{c|}{}                       & \multicolumn{1}{c|}{LL}                  & \multicolumn{1}{c|}{0.94}        & \multicolumn{1}{c|}{0.94}   & \multicolumn{1}{c|}{0.7765}      & \multicolumn{1}{c|}{1.021}       & \multicolumn{1}{c|}{204.5}       & \multicolumn{1}{c|}{23.1}     \\ \cline{2-9}
\multicolumn{1}{|c|}{} &\multicolumn{1}{c|}{\multirow{2}{*}{CW$_0$}} & \multicolumn{1}{c|}{most}                  & \multicolumn{1}{c|}{1}       & \multicolumn{1}{c|}{1}  & \multicolumn{1}{c|}{0.97}        & \multicolumn{1}{c|}{\textbf{2.189}}       & \multicolumn{1}{c|}{59.5}        & \multicolumn{1}{c|}{\textbf{662.7}}    \\ \cline{3-9}
\multicolumn{1}{|c|}{} &\multicolumn{1}{c|}{}                       & \multicolumn{1}{c|}{LL}                  & \multicolumn{1}{c|}{1}       & \multicolumn{1}{c|}{1}  & \multicolumn{1}{c|}{0.8056}      & \multicolumn{1}{c|}{\textbf{3.007}}       & \multicolumn{1}{c|}{207.5}       & \multicolumn{1}{c|}{\textbf{794.9}}    \\ \hline
\end{tabular}
}}
\caption{\small Evaluation of adversarial attacks. }
\label{table:attacks}
\vspace{-0.4cm}
\end{table}

To measure and compare the adverse effect and cost of adversarial examples generated by different attack algorithms, we introduce the following six attack evaluation metrics.

\textbf{Attack Success Rate (ASR)} indicates the percentage of all attack inputs that are successful adversarial examples. An adversarial example is successful when the attack goal is achieved: misclassification in untargeted attacks or reaching the target label in targeted attacks.

\textbf{Misclassification Rate (MR)} is the percentage of adversarial attack inputs that are misclassified. MR equals to ASR in untargeted attacks (UA) and may be larger than ASR in targeted attacks (TA). A targeted attack example may fail being classified as the target label and land on a misclassified label.

\textbf{Mean confidence on adversarial class (AdvConf)} is the mean confidence on the adversarial class of successful adversarial examples: $\frac{1}{n}\sum\nolimits_{i=1}^n p^*$, where $*$ denotes the misclassified class in UA and the target class in TA.

\textbf{Perturbation Distance Cost (Perturb)} is measured by the root mean square deviation: $\sqrt{\frac{||x-x_{adv}||_2}{n}}$ for successful adversarial examples.

\textbf{Perception Distance Cost (Percept)} is the human perception distance metrics~\cite{luo2018towards} on successful adversarial examples.  Let $V_i$ be the pixel value for pixel $i$. $S_i$ be the 9 or 4 (corner) or 6 (edge)-pixel neighbors of pixel $i$ and $\mu$ be their mean value. Then the Standard Deviation of the patch area (PStd) around pixel $i$ is $\sqrt{\frac{\sum \nolimits_{{V_k}\in{S_i}}(V_i-\mu)}{n^2}}$. The sensitivity ($Sen$) of pixel $i$ is then computed as
$Sen(i) = \left\{ \begin{array}{l}
1,PStd\left( i \right) \le 1\\
\frac{1}{{PStd\left( i \right)}},PStd\left( i \right) > 1
\end{array} \right.$. We compute DistPercept by $\frac{1}{n}\sum \nolimits_i {|\delta_{v_i}|} Sen(S_i)$. $\delta_{v_i}$ denotes the perturbation on pixel $i$.

\textbf{Time cost (Time)} measures the average generation time of a single adversarial example and is measured in seconds.

We evaluate eleven adversarial attacks on CIFAR-10 and ImageNet in Table~\ref{table:attacks} using the above metrics. Consider that some attacks are expensive in per-example time cost, we select the first 100 correctly predicted benign examples in the test set to generate attack examples. For FGSM, $\theta$ is set to 0.0156 for CIFAR-10 and 0.0078 for ImageNet, since with such a small $\theta$ it is sufficient to cause a high attack success rate. For BIM, the per iteration $\theta$ is 0.0012 and 0.002, and the maximum  $\theta$ is 0.008 and 0.004 for CIFAR-10 and ImageNet respectively. For CW attacks, the attack confidence is set to 5 for both CIFAR-10 and ImageNet, and the maximum optimization iteration is set to 1000 since this setting is sufficient for CW algorithms to achieve 100\% attack success rate. The adversarial example is fed into the target ML model every 100 iterations of optimization to check if the attack is successful. For JSMA, the maximum distortion of the image is set to 10\% instead of 15\%, the maximum default in paper~\cite{papernot2016limitations}.

\subsection{Existing Defense Methods}

Existing defense methods against adversarial examples can be classified into three broad categories: adversarial training, gradient masking, and input transformation.
%%, and certified defense.

{\bf Adversarial training} defense techniques retrain the attack target model by using both benign training set and adversarial examples generated using known attack algorithms \cite{goodfellow6572explaining,tramer2017ensemble} in order to improve the generalization of the target model in the presence of adversarial attacks. PGD adversarial training with random restart~\cite{madry2017towards} and multi-model-training on adversarial examples \cite{srisakaokul2018muldef} are the two most recent and more advanced adversarial training algorithms. Adversarial training can be seen as giving the target model a set of flu-shots, one per attack algorithm, and thus, adversarial training defense can be effective against known attacks. However, it also suffers from the attack-dependent training and fails to generalize and against other attacks.

%,vijaykeerthy2018hardening

{\bf Gradient masking} defense techniques add an additional layer of training to reduce the sensitivity of a trained model in response to small changes in input data. The main idea is to obfuscate the gradient information or use a near-zero gradient to off-set or minimize the impact of gradient information manipulation performed by an adversary.
\cite{gu2014towards} adds a gradient penalty term in the model training objective, which is a summation of the layer-wise Frobenius norm of the Jacobian matrix. The gradient penalty makes the target model less sensitive to small perturbations in input, but notably reduces the accuracy and learning effectiveness of the target model.
Defensive distillation~\cite{papernot2016distillation} replaces the last layer with a defensive softmax function and a temperature knot to control the extent of distillation after training.
%aiming to use the distilled model to hide gradient information of the target model from an adversary and reduce the model sensitivity to small changes in input. \cite{carlini2016defensive} shows the vulnerability of defensive distillation against a variant of JSMA with a division trick, and \cite{papernot2018sok} show the limitations of defensive distillation for the black box attacks due to the transferability of adversarial examples~\cite{papernot2016transferability}.

% {\bf Input transformation} techniques are employed to the input data  before sending a query input to the prediction model. Two broad categories of input transformation techniques are employed. One is to simply use popular image preprocessing techniques such as binary filters and median smoothing filters such as feature squeezing~\cite{xu2017feature}, and another is to use the DNN autoencoders~\cite{meng2017magnet,gu2014towards} to leverage the latent patterns to detect adversarial inputs.

{\bf Input transformation} applying careful input manipulation techniques  to the input data before sending it to the target prediction model.  Some popular image processing techniques like binary filters and median smoothing filters are employed in~\cite{xu2017feature} to perform feature squeezing on input data.
%It uses a detection threshold obtained by training on adversarial examples generated by each attack algorithm on each given dataset to perform the detection-only defense.
Other input transformation solutions include PixelDefend~\cite{song2017pixeldefend}, information encoding~\cite{buckman2018thermometer} image quilting and total variation minimization~\cite{guo2017countering}, randomness \cite{xie2017mitigating} or region sampling~\cite{cao2017mitigating}.
The dimension reduction techniques include the PCA projection of benign training data with the top $k$ principle axes~\cite{bhagoji2018enhancing} and the DNN denoising autoencoders \cite{meng2017magnet,samangouei2018defense,gu2014towards}, which learns the latent space representation on training data and leverages the latent patterns to detect adversarial inputs.

%osadchy2017no, das2018shield

% {\bf Certified defense} guarantees that for any given input, the classifier’s prediction is constant within some set around the input, often an L$_2$ or L$_{\infty}$ ball. Satisfiability Modulo Theories~\cite{carlini2017provably,katz2017towards,huang2017safety} and mixed integer linear programming~\cite{tjeng2017evaluating,dutta2017output,cheng2017maximum} are popular tools used for certified adversarial defense design. Global Lipschitz constant~\cite{cisse2017parseval,tsuzuku2018lipschitz}, local Lipschitz constant~\cite{hein2017formal}, and relaxation and duality~\cite{raghunathan2018certified,wong2018scaling,raghunathan2018semidefinite,dvijotham2018dual,dvijotham2018training} are also explored to build certification for the defense.
% However, these methods either cannot scale beyond moderate-sized (100,000 activations) networks or scale to solve large machine learning problems like the ImageNet classification task. Many of these methods are also confined to a certain perturbation norm, leaving adversary the attacking space.  Besides, many of the defense requires a modification of some specific network architectures and subsequently involves retraining of the machine learning model. Thus, certified defense is beyond the scope of this paper and can be considered as future work.

{\bf Defense on OOD Inputs.\/} The first out-of-distribution detection method, proposed by the work~\cite{hendrycks2016baseline}, adds small perturbations to the input data by softmax-controlled input preprocessing and is often referred to as the baseline. One way to improve the baseline method is to combine the input noise injection with output temperature scaling by distillation temperature control~\cite{liang2017enhancing}. However, both detection methods rely on a proper setting of the input noise amount parameter and the output temperature parameter, which are dataset-specific. Another proposal utilizes the Mahalanobis distance with respect to the closest class conditional distribution to flag both out-of-distribution examples and adversarial perturbed examples~\cite{lee2018simple}. Other methods include training robust classifier against out-of-distribution data~\cite{lee2017training}, GAN-based detection~\cite{ryu2018out} and semantic representation~\cite{shalev2018out}.

{\bf Limitations of Existing Defenses.\/} (1) Existing defense methods are sensitive to certain magic parameters inherent in their design, such as the percentage of adversarial examples in a batch in adversarial training, the temperature in Defensive Distillation, the detection threshold trained on benign dataset and adversarial inputs in both input transformation and denoising auto-encoder detector defense. Such dataset-specific and/or attack algorithm-specific control parameters make the defense methods non-adaptive~\cite{papernot2018sok}. (2) Existing detection only defenses are either limited to either out-of-distribution inputs or adversarial examples, whereas other defense methods can only defend adversarial examples but not able to detect OOD input. We argue that a defense system is practical if it is attack-independent and can generalize over attack algorithms. A robust defense approach should not depend on finding the attack-specific or dataset-specific control knob (threshold) to distinguish the adversarial inputs from the benign inputs and to detect and flag those out of distribution inputs.
XEnsemble is developed to promote high robustness and high defensibility of the DNN model against both adversarial examples and out-of-distribution inputs.

\section{XEnsemble Solution Approach}

\subsection{Overview}

\label{section3.1}

During the prediction phase, adversarial examples are sent to the prediction APIs to cause misclassification. Our past work~\cite{wei2020adversarial} identified that different DNN models tend to have different gradient information and thus an adversarial example is often misclassified inconsistently across the different models. For untargeted attacks, such adversarial divergence is reflected by different attack destination classes of the same input. For targeted attacks, the same adversarial example often results in different most-likely and least-likely attack target classes under different prediction models. Although adversarial examples are transferable from one model to another~\cite{papernot2016transferability}, the transferability across different DNN models is not consistent, and the transferability is relatively more severe for untargeted attacks but less so for targeted attacks. For the same adversarial example, the probability of being misclassified by different models to the same wrong class label is not high. Meanwhile, different inputs from the same class also demonstrate some inconsistency under a single model: different destination labels in untargeted attacks and different most-likely and least-likely attack labels in targeted attacks~\cite{wei2020adversarial}. For OOD inputs, we observe that the provided wrongly-predicted label, even under noise with the same distribution and different appearance, is different due to the exhibited uncertainty~\cite{lakshminarayanan2017simple}.
Since both adversarial example and OOD input cause misclassification and both have a low probability of being misclassified to the same wrong class, such prediction discrepancy can be utilized to build a robust prediction model to defend the out-of-distribution input.

Motivated by the characterization of adversarial examples~\cite{wei2020adversarial},  we argue that a robust defense should meet the following two design objectives simultaneously: (1) It should provide a uniform defense architecture that can generalize over both attack algorithms and datasets, and be capable of defending against both adversarial examples and out-of-distribution examples simultaneously. (2) It should be capable of distinguishing the adversarial examples that can be repaired from those that cannot, aiming to maximize the auto-repairing effectiveness. It should also be capable of maximizing the detection rate on the out-of-distribution examples to reduce or minimize the misclassification rate.

\begin{figure}[t]
\centering
\includegraphics[scale=.47]{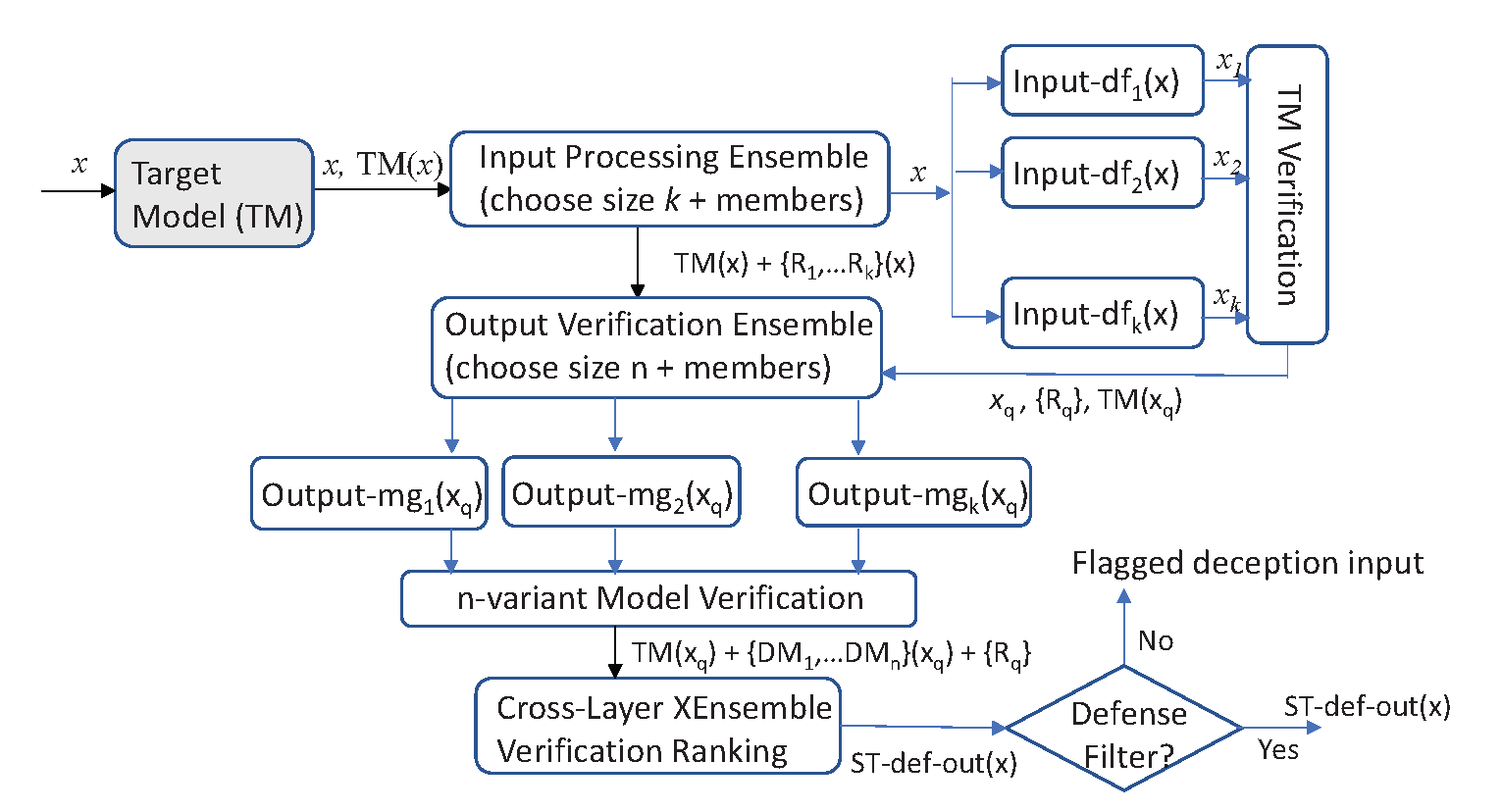}
\caption{{\small Architecture of the XEnsemble defense system}}
\label{figure:architecture}
\end{figure}

We propose to develop XEnsemble, an input-output model verification ensemble defense methodology. XEnsemble is by design unique in two aspects. First, it promotes the composition of disagreement diversity ensembles using high accuracy base models instead of weak models. Second, it combines input verification with output verification using ensemble techniques based on both denoising and failure independent redundancy. XEnsemble can protect a well trained DNN model at the prediction phase without re-training the model using adversarial examples. XEnsemble can be easily added as a plug-in to existing modern neural network architecture.
Figure~\ref{figure:architecture} shows an architectural sketch of the XEnsemble system. It combines diverse input denoising techniques with disagreement diversity output verification techniques. An input $x$ submitted to a DNN model under protection (target model) is first processed by $k$ diverse input denoising techniques, denoted by $input\_df_{1}, input\_df_{2}, \dots, input\_df_{k}$ ($k\geq 3$). We  require to choose only those techniques such that both the target model and the ensemble member models have high test accuracy, on par with their accuracy performance on the original dataset. Second, for each of the $k$ denoising input versions of $x$, say $x_q$, it will be sent to the output-verification ensemble layer, which will select $n$ model verifiers, denote by $output\_mg_{1}, output\_mg_{2}, \dots, output\_mg_{n}$ ($n\geq 3$) for output model verification. XEnsemble will verify, repair and output the defense-approved prediction outcome of the target model, denoted by ST-def-out($x$). The two ensemble size parameters $k$ and $n$ for input denoising process and output model verification are chosen either randomly or by using specific diversity ranking criteria~\cite{yanzhao2020boosting}. Consider output model verification, we first constructed a pool of base models that have diverse network structure or backbone algorithm compared to the target model. This base model pool can be obtained by training new models from scratch or by leveraging the model zoo of pre-trained models for the same task~\cite{xu2017feature}. Then, we compute the ensemble diversity score for each combination of ensemble teams of size 2 or larger from the base model pool. For a base model pool of size $M$=3, 5 or 10, we will have a total of  3, 26, 1013 ensemble teams.  In our first prototype of XEnsemble, we use the $\kappa$ method~\cite{mchugh2012interrater} to compute the $\kappa$ value for each team as the prediction disagreement measure. Third, upon receiving multiple versions of the input $x$ from the input denoising layer, we select an ensemble team from the list of top-ranked ensembles by their ensemble diversity scores, we send each of the $k$ input versions to the chosen diversity ensemble and collect the total of $k\times n$ prediction results. We combine these results by ranking them against a pre-defined confidence level and output the verified prediction results if the degree of the prediction agreement is higher than the pre-defined confidence level, and otherwise, we flag the input (when the agreement is below the threshold). Algorithm~\ref{algo:system} provides a procedural sketch of the XEnsemble method.

\begin{algorithm}[t]
\footnotesize
\caption{\footnotesize  Workflow of XEnsemble}
\begin{algorithmic}[1]
\STATE  \textbf{  Inputs:} \\
$x_0$: query data, $mg_0$: target prediction model, $k$: number of input-denoising modules, $n$: number of output verification models, $\rm T$: $L_1$ ranking confidence level
\\
\vspace{0.1cm}
\STATE \textbf{ensemble preparation}
\STATE select $k$ input-denoising methods $df_1,df_2,...,df_k$ with expertise in processing different kinds of noise.
\STATE generate a base model pool according to the target model and the prediction task
\STATE from the base model pool, create a diversity ensemble model pool using $\kappa$ metric or other diversity metrics.
\STATE given the diversity ensemble model pool, get top $m$ model ensembles with $n$ models: $\{mg_1,mg_2,..,.mg_n\}_{1,...,m}$.
\\
\STATE \textbf{ensemble prediction}
\STATE generate multiple denoised version of querry: $x_1,...,x_k$ from $x_0$ via different noise reduction techniques  $df_1,df_2,...,df_k$.
\STATE feed $x_0,x_1,...,x_k$ to all models in the selected ensembles $\{mg_1,mg_2,..,.mg_n\}_{1,...,m}$ and get ensemble prediction results
\STATE  ranking prediction made by $\{mg_1(x_0),..,.mg_n(x_0)\}_i,...,\{mg_1(x_k),..,.mg_n(x_k)\}_i$, where $i\in \{1,...,m\}$
\IF{$L_1$ distance ranking of the selected ensemble $<$ ranking confidence level}
 \STATE output XEnsemble verified prediction results
\ELSE
\STATE flag the deception input and output a watchout alert.
\ENDIF
\end{algorithmic}
\label{algo:system}
\end{algorithm}

We would like to note that the concept of input denoising and the ensemble diversity are general and applicable to different data modalities, such as image, audio, and text. For example, quantization, filtering, and Gaussian smoothing can be applied to image data; spectral subtraction and Wiener filtering are effective for audio data. In the rest of the paper, we describe in detail our prototype implementation system using the image classification learning as an example application domain. Hence, the input denoising ensemble methods and the output verification ensemble methods are tailored accordingly. We consider three threat scenarios for XEnsemble system when an adversary attacks victim model (target model) based on the level of knowledge about the XEnsemble defense system: (1) zero knowledge (black-box threat model), (2) partial knowledge (grey-box threat model),  and (3) full knowledge (insider adversary) (white box threat model). We first show that when attackers can launch white box attack to the target model but have no prior knowledge of XEnsemble protection. This is the common scenario that is used in the literature for defense proposals. We will also discuss the scenario where attackers have partial (grey box threat) or full knowledge (white box threat) of XEnsemble defense in addition to the target victim model.

\subsection{Input Verification Ensemble}

\begin{table*}[ht]
\centering
\scalebox{0.8}{
\small{
\begin{tabular}{ccccccccccccccc}
\hline
\multicolumn{1}{|c|}{}&
\multicolumn{1}{c|}{\multirow{2}{*}{Attack}} & \multicolumn{1}{c|}{\multirow{2}{*}{}benign} & \multicolumn{1}{c|}{FGSM} & \multicolumn{1}{c|}{BIM} & \multicolumn{1}{c|}{PGD} & \multicolumn{2}{c|}{CW$_\infty$} & \multicolumn{2}{c|}{CW$_2$} & \multicolumn{2}{c|}{CW$_0$} & \multicolumn{2}{c|}{JSMA} & \multicolumn{1}{c|}{\multirow{2}{*}{average}} \\ \cline{4-14}
\multicolumn{1}{|c|}{}&\multicolumn{1}{c|}{} & \multicolumn{1}{c|}{} & \multicolumn{3}{c|}{UA} & \multicolumn{1}{c|}{most} & \multicolumn{1}{c|}{LL} & \multicolumn{1}{c|}{most} & \multicolumn{1}{c|}{LL} & \multicolumn{1}{c|}{most} & \multicolumn{1}{c|}{LL} & \multicolumn{1}{c|}{most} & \multicolumn{1}{c|}{LL} & \multicolumn{1}{c|}{} \\ \hline
\multicolumn{1}{|c|}{\multirow{13}{*}{\rotatebox{90}{CIFAR-10}}}&\multicolumn{1}{c|}{no defense} & \multicolumn{1}{c|}{\textbf{0.9484}} & \multicolumn{1}{c|}{0.15} & \multicolumn{1}{c|}{0.08} & \multicolumn{1}{c|}{0} & \multicolumn{1}{c|}{0} & \multicolumn{1}{c|}{0} & \multicolumn{1}{c|}{0} & \multicolumn{1}{c|}{0} & \multicolumn{1}{c|}{0} & \multicolumn{1}{c|}{0} & \multicolumn{1}{c|}{0} & \multicolumn{1}{c|}{0} & \multicolumn{1}{c|}{0.021} \\ \cline{2-15}
\multicolumn{1}{|c|}{}&\multicolumn{1}{c|}{quan-1-bit} & \multicolumn{1}{c|}{0.2111} & \multicolumn{1}{c|}{0.18} & \multicolumn{1}{c|}{0.17} & \multicolumn{1}{c|}{0.14} & \multicolumn{1}{c|}{0.15} & \multicolumn{1}{c|}{0.17} & \multicolumn{1}{c|}{0.17} & \multicolumn{1}{c|}{0.2} & \multicolumn{1}{c|}{0.16} & \multicolumn{1}{c|}{0.14} & \multicolumn{1}{c|}{0.15} & \multicolumn{1}{c|}{0.16} & \multicolumn{1}{c|}{0.163} \\ \cline{2-15}
\multicolumn{1}{|c|}{}&\multicolumn{1}{c|}{quan-4-bit} & \multicolumn{1}{c|}{0.9311} & \multicolumn{1}{c|}{0.21} & \multicolumn{1}{c|}{0.23} & \multicolumn{1}{c|}{0.17} & \multicolumn{1}{c|}{0.43} & \multicolumn{1}{c|}{0.73} & \multicolumn{1}{c|}{0.54} & \multicolumn{1}{c|}{0.82} & \multicolumn{1}{c|}{0.08} & \multicolumn{1}{c|}{0.08} & \multicolumn{1}{c|}{0.3} & \multicolumn{1}{c|}{0.2} & \multicolumn{1}{c|}{0.345} \\  \cline{2-15}
\multicolumn{1}{|c|}{}&\multicolumn{1}{c|}{medFilter-2*2} & \multicolumn{1}{c|}{0.8929} & \multicolumn{1}{c|}{0.38} & \multicolumn{1}{c|}{0.56} & \multicolumn{1}{c|}{0.40} & \multicolumn{1}{c|}{\textbf{0.73}} & \multicolumn{1}{c|}{0.84} & \multicolumn{1}{c|}{\textbf{0.79}} & \multicolumn{1}{c|}{0.84} & \multicolumn{1}{c|}{\textbf{0.84}} & \multicolumn{1}{c|}{\textbf{0.86}} & \multicolumn{1}{c|}{\textbf{0.89}} & \multicolumn{1}{c|}{0.74} & \multicolumn{1}{c|}{\textbf{0.715}} \\  \cline{2-15}
\multicolumn{1}{|c|}{}&\multicolumn{1}{c|}{medFilter-3*3} & \multicolumn{1}{c|}{0.7502} & \multicolumn{1}{c|}{\textbf{0.42}} & \multicolumn{1}{c|}{\textbf{0.63}} &  \multicolumn{1}{c|}{0.66} & \multicolumn{1}{c|}{0.69} & \multicolumn{1}{c|}{0.69} & \multicolumn{1}{c|}{0.72} & \multicolumn{1}{c|}{0.73} & \multicolumn{1}{c|}{0.77} & \multicolumn{1}{c|}{0.77} & \multicolumn{1}{c|}{0.75} & \multicolumn{1}{c|}{0.75} & \multicolumn{1}{c|}{0.626} \\  \cline{2-15}
\multicolumn{1}{|c|}{}&\multicolumn{1}{c|}{NLM-11-3-2} & \multicolumn{1}{c|}{0.9421} & \multicolumn{1}{c|}{0.2} & \multicolumn{1}{c|}{0.25} & \multicolumn{1}{c|}{\textbf{0.67}} & \multicolumn{1}{c|}{0.36} & \multicolumn{1}{c|}{0.74} & \multicolumn{1}{c|}{0.38} & \multicolumn{1}{c|}{0.78} & \multicolumn{1}{c|}{0} & \multicolumn{1}{c|}{0.01} & \multicolumn{1}{c|}{0.22} & \multicolumn{1}{c|}{0.12} & \multicolumn{1}{c|}{0.339} \\  \cline{2-15}
\multicolumn{1}{|c|}{}&\multicolumn{1}{c|}{NLM-11-3-4} & \multicolumn{1}{c|}{0.9118} & \multicolumn{1}{c|}{0.27} & \multicolumn{1}{c|}{0.46} & \multicolumn{1}{c|}{0.35} & \multicolumn{1}{c|}{0.57} & \multicolumn{1}{c|}{0.85} & \multicolumn{1}{c|}{0.64} & \multicolumn{1}{c|}{0.9} & \multicolumn{1}{c|}{0.05} & \multicolumn{1}{c|}{0.11} & \multicolumn{1}{c|}{0.46} & \multicolumn{1}{c|}{0.32} & \multicolumn{1}{c|}{0.453} \\ \cline{2-15}
\multicolumn{1}{|c|}{}&\multicolumn{1}{c|}{NLM-13-3-2} & \multicolumn{1}{c|}{0.9414} & \multicolumn{1}{c|}{0.19} & \multicolumn{1}{c|}{0.26} & \multicolumn{1}{c|}{0.54} & \multicolumn{1}{c|}{0.37} & \multicolumn{1}{c|}{0.75} & \multicolumn{1}{c|}{0.38} & \multicolumn{1}{c|}{0.81} & \multicolumn{1}{c|}{0} & \multicolumn{1}{c|}{0.01} & \multicolumn{1}{c|}{0.27} & \multicolumn{1}{c|}{0.12} & \multicolumn{1}{c|}{0.336} \\  \cline{2-15}
\multicolumn{1}{|c|}{}&\multicolumn{1}{c|}{NLM-13-3-4} & \multicolumn{1}{c|}{0.9083} & \multicolumn{1}{c|}{0.27} & \multicolumn{1}{c|}{0.45} & \multicolumn{1}{c|}{0.35} & \multicolumn{1}{c|}{0.59} & \multicolumn{1}{c|}{0.85} & \multicolumn{1}{c|}{0.63} & \multicolumn{1}{c|}{0.87} & \multicolumn{1}{c|}{0.05} & \multicolumn{1}{c|}{0.13} & \multicolumn{1}{c|}{0.47} & \multicolumn{1}{c|}{0.32} & \multicolumn{1}{c|}{0.453} \\  \cline{2-15}
\multicolumn{1}{|c|}{}&\multicolumn{1}{c|}{rotation\_-12} & \multicolumn{1}{c|}{0.8335} & \multicolumn{1}{c|}{0.33} & \multicolumn{1}{c|}{0.52} &  \multicolumn{1}{c|}{0.54} & \multicolumn{1}{c|}{0.61} & \multicolumn{1}{c|}{0.73} & \multicolumn{1}{c|}{0.65} & \multicolumn{1}{c|}{0.76} & \multicolumn{1}{c|}{0.32} & \multicolumn{1}{c|}{0.5} & \multicolumn{1}{c|}{0.47} & \multicolumn{1}{c|}{0.34} & \multicolumn{1}{c|}{0.525} \\ \cline{2-15}
\multicolumn{1}{|c|}{}&\multicolumn{1}{c|}{rotation\_-9} & \multicolumn{1}{c|}{0.8529} & \multicolumn{1}{c|}{0.28} & \multicolumn{1}{c|}{0.5} &  \multicolumn{1}{c|}{0.57} & \multicolumn{1}{c|}{0.64} & \multicolumn{1}{c|}{0.79} & \multicolumn{1}{c|}{0.67} & \multicolumn{1}{c|}{0.8} & \multicolumn{1}{c|}{0.27} & \multicolumn{1}{c|}{0.45} & \multicolumn{1}{c|}{0.49} & \multicolumn{1}{c|}{0.36} & \multicolumn{1}{c|}{0.529} \\  \cline{2-15}
\multicolumn{1}{|c|}{}&\multicolumn{1}{c|}{rotation\_3} & \multicolumn{1}{c|}{0.853} & \multicolumn{1}{c|}{0.27} & \multicolumn{1}{c|}{0.33} &  \multicolumn{1}{c|}{0.50} & \multicolumn{1}{c|}{0.55} & \multicolumn{1}{c|}{0.71} & \multicolumn{1}{c|}{0.58} & \multicolumn{1}{c|}{0.7} & \multicolumn{1}{c|}{0.15} & \multicolumn{1}{c|}{0.4} & \multicolumn{1}{c|}{0.36} & \multicolumn{1}{c|}{0.22} & \multicolumn{1}{c|}{0.434} \\  \cline{2-15}
\multicolumn{1}{|c|}{}&\multicolumn{1}{c|}{rotation\_6} & \multicolumn{1}{c|}{0.8493} & \multicolumn{1}{c|}{0.26} & \multicolumn{1}{c|}{0.41} &  \multicolumn{1}{c|}{0.50} & \multicolumn{1}{c|}{0.57} & \multicolumn{1}{c|}{0.7} & \multicolumn{1}{c|}{0.62} & \multicolumn{1}{c|}{0.72} & \multicolumn{1}{c|}{0.22} & \multicolumn{1}{c|}{0.47} & \multicolumn{1}{c|}{0.4} & \multicolumn{1}{c|}{0.3} & \multicolumn{1}{c|}{0.470} \\ \hline
\multicolumn{2}{|c|}{\begin{tabular}[c]{@{}c@{}}med\_2*2, rot\_-12, NLM-13-3-4\end{tabular}} & \multicolumn{1}{c|}{0.8927} & \multicolumn{1}{c|}{0.38} & \multicolumn{1}{c|}{0.60}  & \multicolumn{1}{c|}{\textbf{0.67}} & \multicolumn{1}{c|}{0.72} & \multicolumn{1}{c|}{\textbf{0.89}} & \multicolumn{1}{c|}{0.75} & \multicolumn{1}{c|}{\textbf{0.91}} & \multicolumn{1}{c|}{0.36} & \multicolumn{1}{c|}{0.84} & \multicolumn{1}{c|}{0.67} & \multicolumn{1}{c|}{\textbf{0.76}} & \multicolumn{1}{c|}{0.686} \\ \hline
 \multicolumn{1}{l}{} & \multicolumn{1}{l}{} & \multicolumn{1}{l}{} & \multicolumn{1}{l}{} & \multicolumn{1}{l}{} & \multicolumn{1}{l}{} & \multicolumn{1}{l}{} & \multicolumn{1}{l}{} & \multicolumn{1}{l}{} & \multicolumn{1}{l}{} & \multicolumn{1}{l}{} & \multicolumn{1}{l}{} & \multicolumn{1}{l}{} & \multicolumn{1}{l}{} & \multicolumn{1}{l}{} \\ \hline
\multicolumn{1}{|c|}{\multirow{13}{*}{\rotatebox{90}{ImageNet}}}&\multicolumn{1}{c|}{no defense} & \multicolumn{1}{c|}{0.695} & \multicolumn{1}{c|}{0.01} & \multicolumn{1}{c|}{0}  & \multicolumn{1}{c|}{0} & \multicolumn{1}{c|}{0} & \multicolumn{1}{c|}{0.04} & \multicolumn{1}{c|}{0} & \multicolumn{1}{c|}{0.06} & \multicolumn{1}{c|}{0} & \multicolumn{1}{c|}{0} & \multicolumn{1}{c|}{-} & \multicolumn{1}{c|}{-} & \multicolumn{1}{c|}{0.012} \\ \cline{2-15}
\multicolumn{1}{|c|}{}&\multicolumn{1}{c|}{quan-1-bit} & \multicolumn{1}{c|}{0.24} & \multicolumn{1}{c|}{0.28} & \multicolumn{1}{c|}{0.26} &  \multicolumn{1}{c|}{0.28} & \multicolumn{1}{c|}{0.27} & \multicolumn{1}{c|}{0.28} & \multicolumn{1}{c|}{0.27} & \multicolumn{1}{c|}{0.29} & \multicolumn{1}{c|}{0.28} & \multicolumn{1}{c|}{0.22} & \multicolumn{1}{c|}{-} & \multicolumn{1}{c|}{-} & \multicolumn{1}{c|}{0.270} \\ \cline{2-15}
\multicolumn{1}{|c|}{}&\multicolumn{1}{c|}{quan-4-bit} & \multicolumn{1}{c|}{0.695} & \multicolumn{1}{c|}{0.05} & \multicolumn{1}{c|}{0.05} &  \multicolumn{1}{c|}{0.05} & \multicolumn{1}{c|}{0.31} & \multicolumn{1}{c|}{0.76} & \multicolumn{1}{c|}{0.47} & \multicolumn{1}{c|}{0.82} & \multicolumn{1}{c|}{0.1} & \multicolumn{1}{c|}{0.58} & \multicolumn{1}{c|}{-} & \multicolumn{1}{c|}{-} & \multicolumn{1}{c|}{0.354} \\ \cline{2-15}
\multicolumn{1}{|c|}{}&\multicolumn{1}{c|}{medfilter-2*2} & \multicolumn{1}{c|}{0.65} & \multicolumn{1}{c|}{0.22} & \multicolumn{1}{c|}{0.28} & \multicolumn{1}{c|}{0.26} & \multicolumn{1}{c|}{0.68} & \multicolumn{1}{c|}{0.82} & \multicolumn{1}{c|}{0.74} & \multicolumn{1}{c|}{0.85} & \multicolumn{1}{c|}{0.84} & \multicolumn{1}{c|}{0.85} & \multicolumn{1}{c|}{-} & \multicolumn{1}{c|}{-} & \multicolumn{1}{c|}{0.616} \\ \cline{2-15}
\multicolumn{1}{|c|}{}&\multicolumn{1}{c|}{medfilter-3*3} & \multicolumn{1}{c|}{0.61} & \multicolumn{1}{c|}{0.33} & \multicolumn{1}{c|}{0.41} & \multicolumn{1}{c|}{0.37} & \multicolumn{1}{c|}{0.7} & \multicolumn{1}{c|}{0.8} & \multicolumn{1}{c|}{0.73} & \multicolumn{1}{c|}{0.78} & \multicolumn{1}{c|}{0.79} & \multicolumn{1}{c|}{0.82} & \multicolumn{1}{c|}{-} & \multicolumn{1}{c|}{-} & \multicolumn{1}{c|}{0.637} \\\cline{2-15}
\multicolumn{1}{|c|}{}&\multicolumn{1}{c|}{NLM-11-3-2} & \multicolumn{1}{c|}{0.7} & \multicolumn{1}{c|}{0.05} & \multicolumn{1}{c|}{0.09} & \multicolumn{1}{c|}{0.08} & \multicolumn{1}{c|}{0.28} & \multicolumn{1}{c|}{0.75} & \multicolumn{1}{c|}{0.43} & \multicolumn{1}{c|}{0.87} & \multicolumn{1}{c|}{0.04} & \multicolumn{1}{c|}{0.25} & \multicolumn{1}{c|}{-} & \multicolumn{1}{c|}{-} & \multicolumn{1}{c|}{0.316} \\ \cline{2-15}
\multicolumn{1}{|c|}{}&\multicolumn{1}{c|}{NLM-11-3-4} & \multicolumn{1}{c|}{0.66} & \multicolumn{1}{c|}{0.1} & \multicolumn{1}{c|}{0.25} & \multicolumn{1}{c|}{0.21} & \multicolumn{1}{c|}{0.59} & \multicolumn{1}{c|}{0.83} & \multicolumn{1}{c|}{0.68} & \multicolumn{1}{c|}{0.84} & \multicolumn{1}{c|}{0.2} & \multicolumn{1}{c|}{0.5} & \multicolumn{1}{c|}{-} & \multicolumn{1}{c|}{-} & \multicolumn{1}{c|}{0.467} \\ \cline{2-15}
\multicolumn{1}{|c|}{}&\multicolumn{1}{c|}{NLM-13-3-2} & \multicolumn{1}{c|}{0.7} & \multicolumn{1}{c|}{0.06} & \multicolumn{1}{c|}{0.09} & \multicolumn{1}{c|}{0.09} & \multicolumn{1}{c|}{0.32} & \multicolumn{1}{c|}{0.75} & \multicolumn{1}{c|}{0.46} & \multicolumn{1}{c|}{0.87} & \multicolumn{1}{c|}{0.04} & \multicolumn{1}{c|}{0.27} & \multicolumn{1}{c|}{-} & \multicolumn{1}{c|}{-} & \multicolumn{1}{c|}{0.328} \\ \cline{2-15}
\multicolumn{1}{|c|}{}&\multicolumn{1}{c|}{NLM-13-3-4} & \multicolumn{1}{c|}{0.665} & \multicolumn{1}{c|}{0.11} & \multicolumn{1}{c|}{0.26} &  \multicolumn{1}{c|}{0.23} & \multicolumn{1}{c|}{0.59} & \multicolumn{1}{c|}{0.84} & \multicolumn{1}{c|}{0.68} & \multicolumn{1}{c|}{0.87} & \multicolumn{1}{c|}{0.2} & \multicolumn{1}{c|}{0.51} & \multicolumn{1}{c|}{-} & \multicolumn{1}{c|}{-} & \multicolumn{1}{c|}{0.477} \\ \cline{2-15}
\multicolumn{1}{|c|}{}&\multicolumn{1}{c|}{rotation\_-12} & \multicolumn{1}{c|}{0.62} & \multicolumn{1}{c|}{0.41} & \multicolumn{1}{c|}{0.55} &  \multicolumn{1}{c|}{0.51} & \multicolumn{1}{c|}{0.74} & \multicolumn{1}{c|}{0.74} & \multicolumn{1}{c|}{0.73} & \multicolumn{1}{c|}{0.78} & \multicolumn{1}{c|}{0.68} & \multicolumn{1}{c|}{0.64} & \multicolumn{1}{c|}{-} & \multicolumn{1}{c|}{-} & \multicolumn{1}{c|}{0.642} \\ \cline{2-15}
\multicolumn{1}{|c|}{}&\multicolumn{1}{c|}{rotation\_-9} & \multicolumn{1}{c|}{0.635} & \multicolumn{1}{c|}{0.39} & \multicolumn{1}{c|}{0.53} &  \multicolumn{1}{c|}{0.48} & \multicolumn{1}{c|}{0.77} & \multicolumn{1}{c|}{0.79} & \multicolumn{1}{c|}{0.78} & \multicolumn{1}{c|}{0.8} & \multicolumn{1}{c|}{0.69} & \multicolumn{1}{c|}{0.72} & \multicolumn{1}{c|}{-} & \multicolumn{1}{c|}{-} & \multicolumn{1}{c|}{0.661} \\ \cline{2-15}
\multicolumn{1}{|c|}{}&\multicolumn{1}{c|}{rotation\_3} & \multicolumn{1}{c|}{0.68} & \multicolumn{1}{c|}{0.29} & \multicolumn{1}{c|}{0.44} &  \multicolumn{1}{c|}{0.40} & \multicolumn{1}{c|}{0.71} & \multicolumn{1}{c|}{0.83} & \multicolumn{1}{c|}{0.76} & \multicolumn{1}{c|}{0.82} & \multicolumn{1}{c|}{0.65} & \multicolumn{1}{c|}{0.78} & \multicolumn{1}{c|}{-} & \multicolumn{1}{c|}{-} & \multicolumn{1}{c|}{0.631} \\ \cline{2-15}
\multicolumn{1}{|c|}{}&\multicolumn{1}{c|}{rotation\_6} & \multicolumn{1}{c|}{0.68} & \multicolumn{1}{c|}{0.33} & \multicolumn{1}{c|}{0.49} & \multicolumn{1}{c|}{0.46} & \multicolumn{1}{c|}{0.78} & \multicolumn{1}{c|}{0.85} & \multicolumn{1}{c|}{0.79} & \multicolumn{1}{c|}{0.83} & \multicolumn{1}{c|}{0.72} & \multicolumn{1}{c|}{0.74} & \multicolumn{1}{c|}{-} & \multicolumn{1}{c|}{-} & \multicolumn{1}{c|}{0.666} \\ \hline
\multicolumn{2}{|c|}{\begin{tabular}[c]{@{}c@{}}med-3*3, rot\_-9, rot\_6\end{tabular}} & \multicolumn{1}{c|}{\textbf{0.75}} & \multicolumn{1}{c|}{\textbf{0.53}} & \multicolumn{1}{c|}{\textbf{0.64}}  & \multicolumn{1}{c|}{\textbf{0.65}} & \multicolumn{1}{c|}{\textbf{0.90}} & \multicolumn{1}{c|}{\textbf{0.92}} & \multicolumn{1}{c|}{\textbf{0.89}} & \multicolumn{1}{c|}{\textbf{0.93}} & \multicolumn{1}{c|}{\textbf{0.87}} & \multicolumn{1}{c|}{\textbf{0.89}} & \multicolumn{1}{c|}{-} & \multicolumn{1}{c|}{-} & \multicolumn{1}{c|}{\textbf{0.802}} \\ \hline
\end{tabular}
}}
\caption{\small Defense accuracy of different feature denoising techniques for MNIST, CIFAR-10, and ImageNet}
\label{table:feature_massaging}
\vspace{-0.2cm}
\end{table*}

The main goal of input verification is to apply multiple data modality-specific input noise reduction techniques to clean the input and produce denoised versions of the input, aiming to remove adversarial perturbations or detect the out of distribution inputs and to make the first attempt to prevent adversarial misclassification. While the benign input and their denoised version are prone to be correctly classified, Our work~\cite{wei2020adversarial} shows that the adversarial example and its denoised counterparts will demonstrate prediction discrepancy. While the location of the attack perturbation is carefully chosen and the amount is minimized to meet the imperceptibility, input denoising is applied uniformly on the entire image and is by design to preserve the semantic and visualization quality of the input image. We argue that the input noise reduction techniques should be chosen to preserve certain verifiable properties, such as the test accuracy of the target model on benign inputs, while capable of removing the adverse effect from an adversarial input.

\begin{figure}[t]
\centering
\includegraphics[scale=.68]{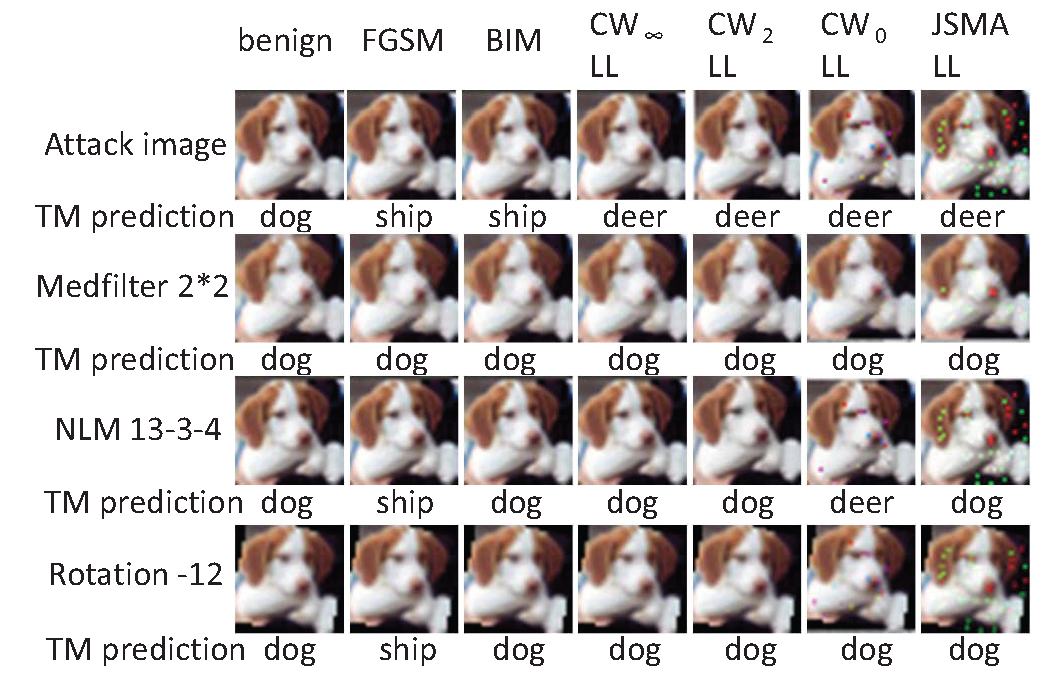}
\vspace{-0.2cm}
\caption{\small CIFAR-10 input-layer defense under 6 attacks.}
\label{figure:cifardefense}
 \vspace{-0.4cm}
\end{figure}

For image classifiers, image smoothing and image augmentation techniques are popular for image noise reduction~\cite{szeliski2010computer}. The former includes pixel quantization by color bit depth reduction, local spatial smoothing, and non-local spatial smoothing. The latter include image rotation, image cropping, and rescaling, image quilting and compression. Although these techniques are not designed to remove the injected noise directly, they can be utilized to make the perturbation less or no longer effective. In the first implementation of XEnsemble, four techniques are implemented using the Scipy library~\cite{SciPy} and OpenCV~\cite{OpenCV-Python}.

\textbf{Rotation} method is a standard image geometric transformation technique provided in SciPy library, rotation preserves the geometric distance of the image and does not change the neighborhood information for most of the pixels except for the corner cases. In our prototype defense, the rotation degree is varied from -12 to 12 with an interval of 3 degree.

The \textbf{Color-depth reduction} is another technique that reduces the color depth of 8 bits ($2^{8}=256$ values) to $i$ bits ($2^{i}$ values). If $i=1$, then the bit quantization will replace the [0,255] space to 1-bit encoding with 2 values: it takes 0 when the nearby pixel value is smaller than 127 and takes 1 when pixel values are in the range of [128, 255]. We use quan-$i$-bit to denote the quantization of the input image from the original 8-bit encoded color depth to $i$-bit ($1\le i <8$).

The \textbf{local spatial smoothing} technique uses nearby pixels to smooth each pixel, with Gaussian, mean or median smoothing. A median filter runs a sliding window over each pixel of the image, where the center pixel is replaced by the median value of the neighboring pixels within the window. The size of the window is a configurable parameter, ranging from 1 up to the image size. MedFilter-$k*k$ denotes the median filter with neighborhood kernel size $k*k$. A square shape window size, e.g., $2\times 2$ or $3\times 3$, is often used with reflect padding.

The \textbf{Non-local spatial smooth (NLM)} technique smooths over similar pixels by exploring a larger neighborhood ($11\times 11$ search window) instead of just nearby pixels and replaces the center patch ( size of $3\times 3$) with the (Gaussian) weighted average of those similar patches in the search window. NLM-$a$-$b$-$c$ denotes non-local means smoothing filter with searching window size $a*a$, patch size $b*b$ and Gaussian distribution parameter $c$. NLM 11-3-4 refers to the NLM filter with $11\times 11$ search window, $3\times 3$ patch size and the filter strength of 4.

%~\cite{szeliski2010computer}~\cite{SciPy-Rotation}~\cite{szeliski2010computer,SciPy-MedianFilter} ~\cite{buades2005non,OpenCV-Python}

Table~\ref{table:feature_massaging} shows the results of input-denoising layer. In this set of experiments, we feed each denoised version to the target model and we show the performance of individual denoising technique and one example denoising ensemble for CIFAR-10 and ImageNet respectively. We make several interesting observations. {\em First}, employing an input noise reduction technique can always improve the robustness of the target model compared to the no-defense scenario. However, there is no single input denoising method that is effective across all 11 attacks. Each technique is good at removing some types of noise but not the other. {\em Second}, not all noise reduction techniques exhibit consistent performance across different datasets trained with different models (e.g., CIFAR-10 with DenseNet and ImageNet with MobileNet as shown in Table~3).
Another interesting observation is that the least-likely attacks are relatively easier to defend than the most-likely attacks. One reason could be that the perturbation in the least-likely attacks is larger and the denoising techniques may work more effectively. In comparison, removing the carefully injected tiny noise in the most-likely attacks may become harder.
Finally, we observe that the input denoising ensemble (see last row of each dataset) show stronger defense compared to individual technique alone, while offering competitive benign test accuracy and providing significantly improved robustness under all 11 attacks. Figure~\ref{figure:cifardefense} shows an intuitive illustration for CIFAR-10.

\begin{figure*}[ht]
\centering
\includegraphics[scale=.81]{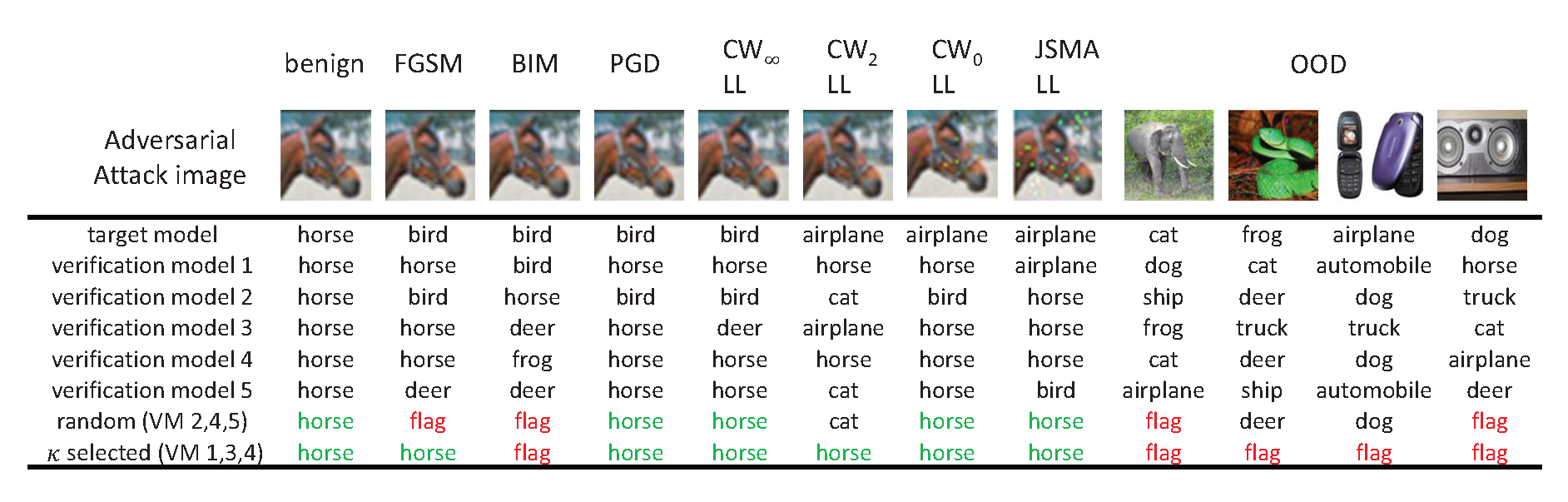}
\vspace{-0.2cm}
\caption{\small An illustration of XEnsemble: model verification ensemble defense against adversarial examples and out-of-distribution attacks.}
\label{figure:defense_illustration}
 \vspace{-0.2cm}
\end{figure*}

\subsection{Output Verification Ensemble}

{\bf Ensemble Diversity.\/}
We define ensemble diversity in terms of both model structure-based diversity and model disagreement-based diversity~\cite{liu2019deep}.
We say that an ensemble meets the structural diversity if the member models of an ensemble are trained using different training process, such as varying training dataset, initial weight filters, neural network structure, neural network algorithm employed (e.g., LeNet, VGG, MobileNet, ResNet-50, ResNet-101, ResNet-152), or by using different settings of hyperparameters (e.g., feature vector size, mini-batch size, \# epochs, \# iterations, learning rate functions, optimization algorithms). Thus, the structural diversity of ensemble is related to the diversity inherent in the algorithms and the hyperparameter settings used for the DNN model training. The disagreement diversity of an ensemble is related to the outputs of the DNN ensemble, which promotes the failure independence of ensemble member models and increases the overall predictive performance (accuracy).
%There are several disagreement measures~\cite{liu2019deep}.

Several popular diversity metrics, e.g., kappa statistics~\cite{mchugh2012interrater}, Q statistic~\cite{huedo2006assessing}, correlation coefficient, fail/non-fail disagreement measure~\cite{sesmero2015generating}), double-fault measure~\cite{kuncheva2003measures} can be used for selecting most disagreement-diverse models from the $N_{out}$ baseline candidate models trained over the same benign training set.
In the first prototype of XEnsemble, we use the Kappa $\kappa$ statistics, a pairwise metric, to measure the diversity degree between two models using negative examples. Let $N$ denote the number of prediction results, $K$ denote the number of classes, $N_{ij}$ denote the number of instances labeled as class $i$ by one model and as class $j$ by the other model. The $\kappa$ metric is defined by
%\begin{align}
\begin{equation}
f_{\kappa} = \frac{{\frac{{\sum\nolimits_{i = 1}^K {{N_{ii}}} }}{N} - \sum\nolimits_{i = 1}^K {(\frac{{{N_{i*}}}}{N} - \frac{{{N_{*i}}}}{N})} }}{{1 - \sum\nolimits_{i = 1}^K {(\frac{{{N_{i*}}}}{N} - \frac{{{N_{*i}}}}{N})} }} \label{equa:kappa}
\end{equation}

%\end{align}
\noindent In Equation~\ref{equa:kappa}, $\frac{{\sum\nolimits_{i = 1}^K {{N_{ii}}}}}{N}$ computes the percentage of agreement made by the two classifiers $i$ and $j$ under the same series of queries. $\sum\nolimits_{i = 1}^K (\frac{{{N_{i*}}}}{N} - \frac{{{N_{*i}}}}{N})$ computes the chance of agreement in which the $*$ is any label in the output space. $\kappa$ value is closer to 1 when two models have more agreements, and is closer to 0 if there is more discrepancy.
The disagreement diversity of an ensemble team of size $M$ is computed by averaging all pairwise $\kappa$ values of its member models.

%Other diversity metrics, such as Q statistic, double-fault measure can also be used.

% ~\cite{huedo2006assessing}
% ~\cite{kuncheva2003measures}

%fail/non-fail disagreement measure~\cite{sesmero2015generating})

{\bf Creating Diversity Ensemble.\/} We develop a two-step diversity ensemble creation algorithm: First, we create a pool of candidate ensemble member models, or so-called base models. Each base model should meet the following two criteria: (1) the structure-based ensemble diversity, and (2) the high benign test accuracy comparable to that of the target model. For benchmark datasets like CIFAR-10, CIFAR-100, ImageNet-1000, instead of training N redundant DNN models, one can also collect those pre-trained DNN models on these benchmark datasets from the public domain. Although generating a pool of base models for datasets in some application domain can be potentially expensive, there are several recent efforts~\cite{lee2016stochastic,huang2017snapshot} present encouraging mechanisms to create multiple base models conveniently, such as within a single training phase using Snapshots.
Second, we need to create disagreement-based ensemble teams from the base model pool. This can be carried out by finding all the subsets of the base model pool of size P, which meet the following two constraints: (i) we want to include the target model in every defense ensemble; and (ii) we want to remove all ensemble teams of size $2$. Thus, instead of considering $2^{P}$ subsets, we only need to consider $2^{P-2}$ subsets to create the pool of disagreement-based diversity ensemble teams from a base model pool of size $P$. We refer to those subsets as the ensemble candidate teams. For each candidate teams, we compute its ensemble disagreement score based on the chosen diversity metric, such as the Kappa $\kappa$ value. Then we rank all $2^{P-2}$ candidate ensemble teams by their $\kappa$ scores and form a pool of $\kappa$ ensembles by selecting the top-ranked ensemble teams whose $\kappa$ values are lower than the system-defined $\kappa$ diversity threshold. The lower the $\kappa$ value of an ensemble, the higher the disagreement-based ensemble diversity.

{\bf Defense with Output Verification Ensemble.\/} The output verification ensemble defense involves two problems: (1) We need to develop robust ensemble consensus methods, which can effectively combine, rank, and integrate predictions from members of an ensemble committee to produce the ensemble prediction output with high accuracy, aiming to improve the robustness of the target model under protection against in adversarial examples and out-of-distribution examples. (2) At prediction phase, for each query input, one ensemble team should be chosen randomly from the disagreement diversity ranked pool of ensemble teams as the chosen output verification team to verify and repair the prediction of the target model by simply providing the ensemble recommendation as the verified output of the target model.

There are several consensus methods for combining the outputs of multiple models in an ensemble of size $M$. For example, an ensemble output can be created by aggregating the outputs from its $M$ member DNN models via simple averaging (or sum, max, median, min) or a weighted averaging method. The weighted averaging embraces the relative accuracy of the ensemble member DNN classifiers, e.g., the confidence of the top-1 class label. In general, averaging and weighted averaging are popular aggregation methods for the linear opinion pools. Voting is a representative non-linear combining method, which combines the individual votes from $M$ member models using rank-based information. The majority voting is the simplest method, which chooses the classification made by more than half of the DNN member classifiers. When there is no agreement among more than 50\% of the DNN member models, the ensemble result is considered an error. The downside of the majority is the scenario where 50\% of an ensemble committee misclassify, and a majority voting, in this case, results in ensemble error. The plurality voting method improves the majority in the sense that the collective decision is the classification reached by more DNN classifiers than any other. A correct decision by the majority is inevitably a correct decision by plurality, but not vice versa~\cite{liu2019deep}.

 \begin{table*}
 \begin{minipage}{0.53\linewidth}
\centering
\scalebox{0.71}{
\small{
 \begin{tabular}{|c|cccccc|}
\hline
 \multirow{2}{*}{CIFAR-10} &  \multicolumn{6}{c|}{$\kappa$ degree of disagreement} \\ \cline{2-7}
  & DenseNet & ResNet-20 & ResNet-32 & ResNet-44 & ResNet-56 & ResNet-110 \\ \hline
TM: DenseNet & 1 & 0.382 & 0.405 & 0.408 & 0.44 & 0.425  \\
VM1: ResNet-20 &  & 1 & 0.496 & 0.435 & 0.452 & 0.478  \\
VM2: ResNet-32 &  &  & 1 & 0.46 & 0.469 & 0.507  \\
VM3: ResNet-44 &  &  &  & 1 & 0.464 & 0.481  \\
VM4: ResNet-56 &  &  &  &  &  & 0.49  \\
VM5: ResNet-110 &  &  &  &  &  & 1  \\ \hline
 benign acc. & 0.9484 & 0.9184 & 0.9228 & 0.9238 & 0.9277 & 0.9262 \\ \hline
\end{tabular}
}}
\subcaption{\small  Kappa statistic for CIFAR-10 models}
\end{minipage}
 \begin{minipage}{0.43\linewidth}
 \vspace{0.2cm}
\centering
\scalebox{0.71}{
\small{
\begin{tabular}{|c|ccccc|}
\hline
 \multirow{2}{*}{ImageNet} &  \multicolumn{5}{c|}{$\kappa$ degree of disagreement} \\ \cline{2-6}
 & MobileNet & VGG-16 & VGG-19 & ResNet-50 & Inception-v3   \\  \hline
TM: MobileNet & 1 & 0.238 & 0.318 & 0.307 & 0.365   \\
VM1: VGG16 &  & 1 & 0.456 & 0.341 & 0.284   \\
VM2: VGG19 &  &  & 1 & 0.33 & 0.285  \\
VM3: ResNet-50 &  &  &  & 1 & 0.399   \\
VM4: Inception-v3 &  &  &  &  & 1   \\ \hline
 benign acc.  & 0.695 & 0.68 & 0.67 & 0.73 & 0.67 \\ \hline
\end{tabular}
}}
\subcaption{\small Kappa statistic for ImageNet models}
\end{minipage}
 \caption{\small Baseline verification model pools for CIFAR-10 and ImageNet with benign test accuracy and pairwise $\kappa$ measures.}
\label{table:kappa_measure}
\end{table*}

We illustrate the model verification ensemble defense against adversarial examples and out-of-distribution attacks in Figure~\ref{figure:defense_illustration}.
%\small
To show the ensemble with only structure diversity and the ensemble with both structure and disagreement diversity, we verify the prediction of the target model using five individual DNN verification models as the pool of base models. We construct ensemble teams of structure diversity by simply choosing a subset of size 3 or larger from the base model pool, which we refer to as the random ensemble team. We also construct a pool of disagreement diversity ensemble teams, which we refer to as the Kappa $\kappa$ ensemble. We highlight those examples that are correctly repaired in green by the random ensemble of a selection of base models and by the Kappa $\kappa$ ranked ensemble team and highlight those that are detected as deceptive inputs in red, which are either OOD inputs or adversarial examples that cannot be repaired by the chosen ensemble team. We observe that the random ensemble is less diverse compared to the Kappa $\kappa$ ranked ensemble team. For instance, two verifiers out of the random ensemble of three base models (Random (VMs 2,4,5) have made misclassification consistently into the same wrong class (VM2 and VM5), e.g., the adversarial input under CW$_2$ attack is misclassified into cat, and the two OOD inputs (Frog and cellphone) are misclassified into deer and dog respectively.
This comparison also shows the importance of developing a comprehensive ensemble ranking algorithm that takes into account not only the prediction outcome but also other factors, such as the prediction confidence, the $L_1$ distance of the verification vector to the target prediction vector. A preliminary version of XEnsemble on the robustness of adversarial example is provided in our conference version~\cite{wei2020cross}.

\section{Experimental Evaluation}

\subsection{Output Verification Implementation}

%The proposed defense is built on top of Evade-ML~\cite{xu2017feature}.

% \cite{deng2009imagenet}
%,\cite{yu2015lsun}

{\bf Datasets.\/} Two popular benchmark datasets CIFAR-10 and ImageNet are used for all experiments reported in this paper.  The CIFAR-10 dataset consists of 60,000 color images of objects with a size of 32*32 in 10 classes. 50,000 of them are training data and the rest 10,000 are test data. The ImageNet dataset is provided by the ImageNet Large Scale Visual Recognition Competition (ILSVRC) as a large-scale color image classification benchmark. The dataset contains 1.2 million training images and 50,000 validation images in 1000 classes with a size of 224*224.  We also use the TinyImageNet data and LSUN data as the two out-of-distribution datasets, which are fed into a DenseNet classifier trained on CIFAR-10 and CIFAR-100 for the effectiveness evaluation and comparative analysis. CIFAR-100 dataset consists of 50,000 32x32 color training images, labeled with 100 classes, and 10,000 test images. TinyImageNet is a subset of the ImageNet dataset, containing 10,000 test images from 200 different classes. The Large-scale Scene UNderstanding dataset (LSUN) has a test set of 10,000 images on 10 different scenes.

%~\cite{huang2017densely}

{\bf Creating Base Model Pool.\/} For CIFAR-10 trained on DenseNet with test accuracy of 0.9184, we create a base model pool of 5 pre-trained models as candidate verifiers for the target DenseNet model: ResNet-20, ResNet-32, ResNet-44, ResNet-56, and ResNet-110, all obtained from Keras. For ImageNet trained on MobileNet with test accuracy of 0.67, we utilize 4 pre-trained models from Keras as the verification models (VMs) for the target MobileNet model: VGG-16, VGG-19, ResNet-50, and Inception-V3. In addition to the structure-based ensemble diversity, we also use the benign test accuracy as another criterion for our base model selection decision, which requires each model verifier is trained on the same dataset with the same learning task and has the benign test accuracy comparable to that of the target model. Table~\ref{table:kappa_measure} shows that all base verification models for CIFAR-10 models have a benign accuracy higher than 0.9184 and all ImageNet baseline verification models have a benign accuracy higher than 0.67.

\begin{table}[t]
\centering
\scalebox{0.73}{
\small{
\begin{tabular}{|c|c|c|c|c|c|c|c|c|}
\hline
\multicolumn{5}{|c|}{CIFAR-10} & \multicolumn{4}{c|}{ImageNet} \\ \hline
\#model & 2 & 3 & 4 & 5 & \#model & 2 & 3 & 4 \\ \hline
top 1 & 1,3 & 1,3,4 & 1,3,4,5 & 1-5 & top 1 & 1,3 & 1,3,4 & 1-4 \\
$\kappa$ & 0.408 & 0.43 & 0.441 & 0.453 & $\kappa$ & 0.295 & 0.322 & 0.332 \\
acc & 0.9476 & 0.9604 & 0.9365 & 0.95 & acc & 0.775 & 0.755 & 0.805 \\ \hline
top 2 & 2,3 & 1,2,3 & 1,3,4,5 & - & top 2 & 1,4 & 1,2,4 & - \\
$\kappa$ & 0.424 & 0.431 & 0.446 & - & $\kappa$ & 0.296 & 0.324 & - \\
acc & 0.9285 & 0.9324 & 0.9301 & - & acc & 0.77 & 0.75 & - \\ \hline
top 3 & 1,4 & 1,3,5 & 1,2,3,5 & - & top 3 & 2,3 & 1,2,3 & - \\
$\kappa$ & 0.428 & 0.435 & 0.448 & - & $\kappa$ & 0.318 & 0.332 & - \\
acc & 0.9198 & 0.8928 & 0.9256 & - & acc & 0.75 & 0.745 & - \\ \hline
\end{tabular}
}}
\caption{\small $\kappa$ value and the benign ensemble accuracy of Top 3 ranked $\kappa$ ensemble teams of different sizes for CIFAR-10 and ImageNet, }
\label{table:kappa_list}
\vspace{-0.2cm}
\end{table}

{\bf Creating Diversity Ensemble Pool.\/} Next, we create a pool of disagreement-based diversity ensembles by pruning those ensembles with low disagreement diversity. In the first prototype of XEnsemble, we quantify the ensemble disagreement diversity using Kappa($\kappa$) measure. Given that Kappa($\kappa$) measure is a pairwise method, we compute the Kappa($\kappa$) score for a given ensemble of size $M$ by computing the average Kappa score, denoted by $\kappa$-average, over all pairwise Kappa($\kappa$) scores of an ensemble.
We identify the best Kappa($\kappa$) ensemble team by selecting the top-1 ranked ensemble in the pool of disagreement ensemble. There are different ways to create a pool of disagreement based verification ensembles. Alternative to the approach discussed in Section 3.2, we can build this pool by creating multiple $\kappa$ ranked ensemble list, one for each size of the ensembles such as 3, 4, 5, 6 for the target DenseNet model (TM) on CIFAR-10 and the base model pool of 5 different Resnet model verifiers. For each Kappa ($\kappa$) ranked ensemble list of a given size, we add the verification ensemble with the smallest average pairwise Kappa ($\kappa$) value to the disagreement based ensemble pool. Table~\ref{table:kappa_list} shows the pool of top-3 Kappa ($\kappa$) ensembles with ensemble prediction accuracy and average pairwise $\kappa$ value for both DenseNet on CIFAR-10 and MobileNet on ImageNet-1000.

{\bf Ensemble Ranking.\/}
There are several ways to perform prediction ensemble ranking. In our first prototype implementation, we choose to use the $L_{1}$ distance and the prediction confidence as two metrics to design our ranking algorithm. We choose the $L_{1}$ distance because we only consider those verification models with high benign test accuracy. Then, the predictions for a benign example on both the target model and verification models should be alike. Thus, the $L_{1}$ distance of their prediction vectors should be small and close to zero. At the same time, the predictions for an adversarial example are likely to be quite different on the target model and verification model. This is feasible since the transferability on targeted attacks is weak. In the meantime, transferability on untargeted attacks merely causes misclassification and the misclassified destination different across prediction models. Therefore, the $L_{1}$ distance between prediction made by the target model and by the verification is relatively larger (higher than 1 and close to 2). Consider $k$ verification versions to the target model (TM), we produce a $L_{1}$ ranking of all $k$ prediction outcomes. If the top-ranked recommendations have approximately the same $L_{1}$ ranking score, we use the prediction confidence for each of the prediction recommendations to break the tie.

\subsection{Defense Evaluation Metrics}

We evaluate the effectiveness of our model verification ensemble defense approach and compare it with other existing defenses by a number of cost and effect metrics.

\textbf{Defense Success Rate (DSR).\/} DSR measures the success rate of the defense system by verifying, repairing and flagging adversarial examples and out-of-distribution examples.

\textbf{Prevention Success Rate (PSR).\/} PSR measures the percentage of adversarial examples and out-of-distribution examples that are repaired and correctly classified by the defense.

\textbf{Detection Success Rate (TSR).\/} TSR measures the percentage of adversarial examples and out-of-distribution examples that cannot be repaired but are flagged by the defense.

When adversarial examples and out-of-distribution examples are presented as threats to a target DNN model, the defense success rate (DSR) is the sum of PSR and TSR, i.e., DSR = PSR + TSR. Since there is no correct label for the out-of-distribution input in a trained model, the PSR for OOD is 0 and DSR equals to TSR. To better measure the detection performance of XEnsemble, we also use the following common metrics to evaluate the effectiveness of model verification ensemble defense for detecting out-of-distribution examples in addition to TSR.

\textbf{Detection error (DError)} measures the error made both on out-of-distribution and in-distribution data. It is computed by $\beta$(1-TPR)+(1-$\beta$)FPR. TPR is computed by TP / (TP+FN), where TP denotes the portion of correctly detected out-of-distribution examples and FN denotes the portion of OOD inputs that are considered as in-distribution data. FPR is computed by FP/(FP+TN), where FP is the proportion of in-distribution inputs that are identified as out-of-distribution inputs and TN is the proportion of in-distribution data that are identified correctly. We set $\beta=0.5$.
%due to an equal number of out-of-distribution and in-distribution data in our experiment.

\textbf{AUROC} is a threshold-independent metric that measures the tradeoff between (1-FPR) and TPR. The range of AUROC is [0,1]. It can be interpreted as the probability that a positive example is assigned a higher detection score than a negative example. The better the detection performance is, the larger the AUROC will be.

%~\cite{davis2006relationship}

\begin{table*}[ht]
\centering
\scalebox{0.80}{
\small{
\begin{tabular}{|c|c|c|c|c|c|c|c|c|c|c|c|c|c|c|}
\hline
\multirow{11}{*}{\rotatebox{90}{CIFAR-10}} & \multirow{2}{*}{combo} & \multirow{2}{*}{benign} & \multirow{1}{*}{FGSM$_{\infty}$} & \multirow{1}{*}{BIM$_{\infty}$} &  \multirow{1}{*}{PGD$_{\infty}$} & \multicolumn{2}{c}{CW$_\infty$} & \multicolumn{2}{c}{CW$_2$} & \multicolumn{2}{c}{CW$_0$} & \multicolumn{2}{c|}{JSMA$_{0}$} & \multirow{2}{*}{average} \\ \cline{4-14}
  &  &  &  \multicolumn{3}{c|}{UA} & most & LL & most & LL & most & LL & most & LL &  \\ \cline{2-15}
 & TM(no defense) & 0.9484 & 0.15 & 0.08 & 0.14 & 0 & 0 & 0 & 0 & 0 & 0 & 0 & 0 & 0.034 \\ %\cline{3-15}
%& BagEns & 0.8415 & 0.65 & 0.64 & 0.83 & 0.91 & 0.89 & 0.87 & 0.64 & 0.67 & 0.72 & 0.53 & 0.733\\
%& TM  & 0.945 & 0.15 & 0.08 & 0 & 0  & 0 & 0  & 0    & 0   & 0   & 0   &  0.023 \\
& VM1  & 0.918 & 0.64 & 0.77 & 0.86 & 0.79 & 0.76 & 0.8 & 0.83 & 0.59 & 0.53 & 0.6 & 0.33 & 0.682 \\
& VM2 & 0.923 & 0.6 & 0.75 & 0.84 & 0.76 & 0.73 & 0.8 & 0.78 & 0.66 & 0.56 & 0.58 & 0.35 & 0.674 \\
& VM3  & 0.924 & 0.59 & 0.73 & 0.83 & 0.73 & 0.75 & 0.77 & 0.77 & 0.56 & 0.5 & 0.61 & 0.35 & 0.648 \\
& VM4 & 0.928 & 0.59 & 0.7 & 0.81& 0.77 & 0.73 & 0.81 & 0.74 & 0.59 & 0.42 & 0.56 & 0.32 & 0.64 \\
& VM5 & 0.926 & 0.62 & 0.72  & 0.84& 0.78 & 0.77 & 0.82 & 0.79 & 0.59 & 0.49 & 0.62 & 0.39 & 0.675 \\ \cline{2-15}
& \textbf{XEnsemble-rand} & 0.8928 & 0.71 & 0.82 & 0.86 & 0.9 & 0.88 & 0.91 & 0.91 & 0.72 & 0.69 & 0.8 & 0.56 & 0.796 \\ %\cline{3-15}
& \textbf{XEnsemble-$\kappa$-rand} & 0.95 & \textbf{0.76} & \textbf{0.88} & 0.93 &  0.91 & 0.91 & 0.9 & 0.93 & \textbf{0.73} & 0.79 & 0.82 & 0.65 & 0.837 \\ %\cline{3-15}
& \textbf{XEnsemble-best-$\kappa$} & \textbf{0.9604} & 0.73 & 0.85 & 0.91& \textbf{0.97} & \textbf{0.95} & \textbf{0.97} & \textbf{0.94} & \textbf{0.73} & \textbf{0.87} & \textbf{0.91} & \textbf{0.75} & \textbf{0.871} \\ \hline %\cline{2-14}
\end{tabular}
}}
\caption{\small Measurement of the Defense Success Rate on base models and XEnsemble on CIFAR-10}
\label{table:defense_table_cifar10}
\end{table*}

\begin{table*}[ht]
\centering
\scalebox{0.80}{
\small{
\begin{tabular}{|c|c|c|c|c|c|c|c|c|l}
\cline{1-9}
\multicolumn{2}{|c|}{\multirow{3}{*}{CIFAR-10}} & \multirow{3}{*}{\begin{tabular}[c]{@{}c@{}}test \\ acc\end{tabular}} & \multicolumn{3}{c|}{output verification ensemble (DSR/PSR/TSR)} & \multicolumn{3}{c|}{input-output verification ensemble (DSR/PSR/TSR)} &  \\ \cline{4-9}
\multicolumn{2}{|c|}{} &  & XEnsemble-rand & XEnsemble-$\kappa$-rand & XEnsemble-best-$\kappa$ & XEnsemble-rand & XEnsemble-$\kappa$-rand & XEnsemble-best-$\kappa$ &  \\ \cline{4-9}
\multicolumn{2}{|c|}{} &  & VM 1,3,5 & VM 1-5 & VM 1,3,4 & VM 1,3,5 & VM 1-5 & VM 1,3,4 &  \\ \cline{1-9}
\multicolumn{2}{|c|}{no attack} & 0.9484 & 0.8928/0.8775/0.0153 & 0.95/0.9412/0.0088 & 0.9604/0.9412/0.0192 & 0.9186/0.8826/0.036 & 0.9253/0.9079/0.0174 & 0.9446/8884/0.9446 &  \\ \cline{1-9}
\multicolumn{2}{|c|}{FGSM} & 0.15 & 0.71/0.68/0.03 & 0.76/0.69/0.07 & 0.72/0.69/0.04 & 0.85/0.73/0.12 & 0.91/0.77/0.14 & 0.93/0.78/0.15 &  \\ \cline{1-9}
\multicolumn{2}{|c|}{BIM} & 0.08 & 0.82/0.76/0.05 & 0.88/0.85/0.03 & 0.85/0.85/0 & 0.91/0.84/0.07 & 0.94/0.82/0.12 & 0.99/0.85/0.14 &  \\ \cline{1-9}
\multicolumn{2}{|c|}{PGD} & 0.14 & 0.86/0.81/0.05 & 0.93/0.90/0.03 & 0.91/0.87/0.04 & 0.93/0.88/0.05 & 0.93/0.86/0.07 & 0.95/0.89/0.06 &  \\ \cline{1-9}
\multirow{2}{*}{CW$_\infty$} & most & 0 & 0.90/0.87/0.03 & 0.91/0.87/0.04 & 0.97/0.96/0.01 & 0.92/0.90/0.02 & 0.94/0.86/0.08 & 0.97/0.88/0.09 &  \\ \cline{2-9}
 & LL & 0 & 0.88/0.84/0.04 & 0.91/0.88/0.03 & 0.95/0.93/0.02 & 0.92/0.90/0.02 & 0.96/0.89/0.07 & 0.98/0.88/0.10 &  \\ \cline{1-9}
\multirow{2}{*}{CW$_2$} & most & 0 & 0.91/0.88/0.03 & 0.90/0.85/0.05 & 0.97/0.94/0.03 & 0.92/0.88/0.02 & 0.92/0.85/0.07 & 0.97/0.89/0.08 &  \\ \cline{2-9}
 & LL & 0 & 0.91/0.87/0.04 & 0.93/0.87/0.06 & 0.94/0.93/0.01 & 0.99/0.90/0.09 & 0.98/0.86/0.12 & 0.99/0.91/0.08 &  \\ \cline{1-9}
\multirow{2}{*}{CW$_0$} & most & 0 & 0.72/0.68/0.04 & 0.73/0.68/0.05 & 0.73/0.70/0.03 & 0.97/0.93/0.04 & 0.94/0.87/0.07 & 0.98/0.92/0.06 &  \\ \cline{2-9}
 & LL & 0 & 0.69/0.64/0.05 & 0.79/0.73/0.06 & 0.87/0.78/0.09 & 0.98/0.91/0.07 & 0.94/0.81/0.13 & 1/0.91/0.09 &  \\ \cline{1-9}
\multirow{2}{*}{JSMA} & most & 0 & 0.8/0.74/0.06 & 0.82/0.75/0.07 & 0.91/0.75/0.16 & 0.92/0.90/0.02 & 0.96/0.83/0.13 & 0.97/0.90/0.07 &  \\ \cline{2-9}
 & LL & 0 & 0.56/0.44/0.12 & 0.65/0.59/0.06 & 0.75/0.59/0.16 & 0.88/0.72/0.16 & 0.91/0.75/0.16 & 0.95/0.80/0.15 &  \\ \cline{1-9}
\end{tabular}
}}
\caption{\small  Defense success measurements in DSR/PSR/TSR for alternative XEnsemble algorithms on CIFAR-10}
\label{table:defense_detail_cifar10}
\end{table*}

\begin{table}[ht]
\centering
\scalebox{0.80}{
\small{
\begin{tabular}{|c|c|c|c|c|c|c|c|c|c|}
\hline
\multicolumn{2}{|c|}{attack\textbackslash{}model} & TM  & VM 1 & VM 2 & VM 3 & VM 4 & VM 5 & Best$\kappa$ \\ \hline
FGSM                         & \multirow{3}{*}{UA} & 1 & 0.353 & 0.412 & 0.424 & 0.436 & 0.424 & 0.235 \\ \cline{1-1} \cline{3-9}
BIM                          &                     & 1  & 0.196 & 0.217 & 0.25 & 0.283 & 0.261 & 0.120 \\ \cline{1-1} \cline{3-9}
PGD                          &                     & 1  & 0.116 & 0.128 & 0.139 & 0.174 & 0.139 & 0.066 \\ \cline{1-1} \cline{2-9}
\multirow{2}{*}{CW$_\infty$} & most                  & 1  & 0.09   & 0.11  & 0.12  & 0.15  & 0.13 & 0 \\ \cline{2-9}
                             & LL                  & 1  & 0.02  & 0.01  & 0.02  & 0.03  & 0.01 & 0 \\ \hline
\multirow{2}{*}{CW$_2$}      & most                  & 1 & 0.1   & 0.1   & 0.10  & 0.11   & 0.1 & 0  \\ \cline{2-9}
                             & LL                  & 1  & 0.01  & 0.01  & 0.02  & 0.03  & 0 & 0    \\ \hline
\multirow{2}{*}{CW$_0$}      & most                  & 1 & 0.3   & 0.23  & 0.23  & 0.25  & 0.24 & 0.09  \\ \cline{2-9}
                             & LL                  & 1   & 0.08  & 0.10  & 0.06  & 0.12   & 0.09 & 0 \\ \hline
\multirow{2}{*}{JSMA}        & most                  & 1 & 0.10  & 0.10  & 0.09  & 0.13   & 0.1  & 0 \\ \cline{2-9}
                             & LL                  & 1   & 0.03  & 0.05  & 0.03  & 0.05  & 0.02 & 0 \\ \hline
\multicolumn{2}{|c|}{model average} & 1 & 0.127 & 0.133 & 0.135 & 0.160 & 0.132 & 0.046 \\ \hline
\end{tabular}
}}
\caption{\small Transferability of adversarial examples to five CIFAR-10 verification models and the best Kappa ensemble team. }
\label{table:cifar_asr_allmodel}
\vspace{-0.4cm}
\end{table}

\begin{table*}[ht]
\centering
\scalebox{0.80}{
\small{
\begin{tabular}{|c|c|c|c|c|c|c|c|c|c|c|c|c|c|c|c|c|}
\hline
 \multirow{2}{*}{CIFAR-10} & \multirow{2}{*}{defense} & \multirow{2}{*}{benign} & \multirow{1}{*}{FGSM$_{\infty}$} & \multirow{1}{*}{BIM$_{\infty}$} &  \multirow{1}{*}{PGD$_{\infty}$} & \multicolumn{2}{c|}{CW$_\infty$} & \multicolumn{2}{c|}{CW$_2$} & \multicolumn{2}{c|}{CW$_0$} & \multicolumn{2}{c|}{JSMA$_{0}$} & \multicolumn{2}{c|}{\multirow{1}{*}{\begin{tabular}[c]{@{}c@{}}OOD\end{tabular}}} &
 \multirow{2}{*}{\begin{tabular}[c]{@{}c@{}}avg \\ DSR\end{tabular}}\\ \cline{4-16}
  &  &  &  \multicolumn{3}{c|}{UA} & most & LL & most & LL & most & LL & most & LL & \footnotesize{TinyImageNet} & LSUN & \\ \hline
 \multirow{1}{*}{no defense} & TM & 0.9484 & 0.15 & 0.08 & 0.14 & 0 & 0 & 0 & 0 & 0 & 0 & 0 & 0 & 0 & 0 & 0.028 \\ \cline{1-17}
 \multirow{5}{*}{\begin{tabular}[c]{@{}c@{}}Defense \\ with no \\ detection\end{tabular}} & Advtrain & 0.879 & 0.64 & 0.58 & 0.63 & 0.68 & 0.77 & 0.75 & 0.79 & 0.44 & 0.48 & 0.5 & 0.45 & 0 & 0 & 0.516 \\
 & EnsTrain & 0.8985 & 0.69 & 0.61 & 0.67 & 0.71 & 0.80 & 0.81 & 0.82 & 0.53 & 0.59 & 0.60 & 0.59 & 0 & 0 & 0.571 \\
 & PGD-defense & 0.8675  & \textbf{0.76} & \textbf{0.68} & \textbf{0.86} & 0.76 & 0.84 & 0.78 & 0.86 & 0.55 & 0.59 & 0.61 & \textbf{0.64} & 0 & 0 & \textbf{0.61}  \\

& DefDistill & \textbf{0.9118} & 0.6 & 0.65 & 0.64 & \textbf{0.79} & \textbf{0.88} & \textbf{0.86} & \textbf{0.90} & \textbf{0.60} & \textbf{0.69} & \textbf{0.70} & 0.47 & 0 & 0 & 0.598 \\
& EnsTrans & 0.8014 & 0.23 & 0.4 & 0.51 & 0.56 & 0.61 & 0.57 & 0.61 & 0.19 & 0.34 & 0.45 & 0.41 & 0 & 0 & 0.375 \\ \cline{1-17}
%& EnsBagging &  &  &  &  &  &  &  &  &  &  &  &  \\ \hline

 \multirow{4}{*}{\begin{tabular}[c]{@{}c@{}}Defense \\ with only \\ detection\end{tabular}} & FS & 0.894 & 0.133 & 0.55 & 0.667 & 0.64 & 1 & 0.83 & 1 & 0.73 & 1 & 0.041 & 0.298 & 0.802 & 0.774 & 0.651 \\
& MagNet & 0.873 & 0.92 & \textbf{1} & \textbf{1} & 0.94 & \textbf{1} & \textbf{0.97} & \textbf{1} & 0.75 & 0.95 & 0.85 & 0.96  & \textbf{0.928} & \textbf{0.968} & 0.941 \\
& NIC  & 0.912  & \textbf{1} & \textbf{1} & \textbf{1} &  \textbf{1} & \textbf{1} & 0.94 & \textbf{1} & \textbf{0.95} & \textbf{1} & \textbf{0.90} & \textbf{0.99} & 0.905 & 0.921 & \textbf{0.969} \\
& LID & \textbf{0.944} & 0.53 & 0.52 & 0.69 & 0.69 & 0.84 & 0.67 & 0.85 & 0.73 & 0.82 & 0.76 & 0.69 & 0.877 & 0.949 & 0.733 \\ \cline{1-17}

 \multirow{3}{*}{\begin{tabular}[c]{@{}c@{}}Prevention \\ and \\ detection\end{tabular}}& XEnsemble-rand & 0.9186 & 0.85 & 0.91 &  \textbf{0.96} & 0.92 & 0.92 & 0.92 & \textbf{0.99} & 0.97 & 0.98 & 0.92 & 0.88 & 0.996 & 0.986 & 0.939 \\ %\cline{3-15}
& XEnsemble-$\kappa$-rand & 0.9253 & 0.91 & 0.94 & 0.93 &  0.94 & 0.96 & 0.92 & 0.98 & 0.94 & 0.94 & 0.96 & 0.91 & \textbf{0.998} & 0.992 & 0.943  \\ %\cline{3-15}
& XEnsemble-best-$\kappa$ & \textbf{0.9446} & \textbf{0.93} & \textbf{0.99} & 0.95 & \textbf{0.97} & \textbf{0.98} & \textbf{0.97} & \textbf{0.99} & \textbf{0.98} & \textbf{1} & \textbf{0.97} & \textbf{0.95} & \textbf{0.998} & \textbf{0.996} & \textbf{0.975} \\ \hline %\cline{2-14}
\end{tabular}
}}
\caption{\small Comparison of XEnsemble with existing defenses on CIFAR-10. DSR=PSR+TSR, TSR=0 for the defenses with no detection and PSR=0 for the defenses with only detection (highlight the best in bold for each category}
\label{table:comparison_table}
\end{table*}

% \section{Experimental Evaluation}

\subsection{Defending Adversarial Attacks on CIFAR-10}

We evaluate the effectiveness of XEnsemble defense for verifying and repairing the prediction decision made by the target model in the presence of adversarial examples. This set of experiments compares three different model verification ensemble defense options: XEnsemble-rand, XEnsemble-$\kappa$-rand, and XEnsemble-best-$\kappa$. XEnsemble-rand is an ensemble created from the base model pool with structure-based diversity. XEnsemble-$\kappa$-rand is an ensemble with disagreement diversity, selected randomly from the Kappa ranked list of ensembles whose Kappa ensemble diversity measure is above the system-defined diversity threshold. XEnsemble-best-$\kappa$ is the ensemble with the highest diversity in the Kappa $\kappa$ ranked list of ensembles. For CIFAR-10, the XEnsemble-best-$\kappa$ ensemble is VM 1,3,4. We provide the defense performance measurement in Table~\ref{table:defense_table_cifar10}. We observe three interesting facts: (1) XEnsemble strategy is always beneficial
regardless whether it is the random ensemble from the
baseline verification model pool, or the random $\kappa$ ensemble
from the pool of $\kappa$ ranked ensemble list, or the best $\kappa$ ensemble.
(2) the best $\kappa$ model ensemble is the most effective over
all 11 attacks in terms of average defense success rate (prediction
accuracy). (3) When defending adversarial example,
the model verification ensemble defense is capable of repairing
most of the adversarial perturbed input with high PSR.
In particular, as shown in Table~\ref{table:defense_detail_cifar10}, DSR is further improved by TSR when the verification
cannot agree with each other.

{\bf  Improvement with Input-Model Ensemble.\/} In Table~\ref{table:defense_detail_cifar10}, we also show the improvement of defense performance when we combine the output model verification layer with the input defense layer and build the input-model verification diversity ensemble.
In our prototype, we apply a median filter with window size 2*2, a rotation with -12 degree, and a non-local mean filter NLM 13-3-4 to the CIFAR-10 input and then feed the processed input to multiple models. There can be other ways to combine the diverse input processing techniques with diverse output models and we consider it as future work. The input-model verification XEnsembles improves the model XEnsemble defense significantly.

{\bf Effect of Transferability.\/} The transferability of adversarial examples is widely reported in literature~\cite{papernot2016transferability}. An immediate question one would ask is how the XEnsemble of multiple model verifiers will cope with the transferability of adversarial examples.
Due to the limitation of space, we report the attack transferability experiments only on the CIFAR-10 dataset. For CIFAR-10, we measured the transferability of adversarial examples from the target model (TM) on all 7 defense models and the Best $\kappa$  ensemble (TM, VM 1,3,4) (recall Table~\ref{table:defense_detail_cifar10}).
The results are shown in Table~\ref{table:cifar_asr_allmodel}. In the ensemble, we measure transferability basing on the following criteria. For untargeted attacks, we consider cases where multiple models agree on the same wrong prediction as untargeted transferability since strong prediction discrepancy will be captured by the proposed defense and the query will subsequently be flagged. For targeted attacks, we consider an adversarial example is transferable only when it is being misclassified to the same target label by other models.

We observe three interesting facts. (1) The untargeted attack (by FGSM or BIM) is less effectively mitigated if we only use one defense model due to transferability. (2) For all targeted attacks, one additional DM model can reasonably reduce the transferability. (3) The transferability to the Best$\kappa$ ensemble becomes very small and mostly negligible. Combining the defense success rate result for the Best$\kappa$ ensemble in Table~\ref{table:defense_table_cifar10} and the transferability results for the Best$\kappa$ ensemble in Table~\ref{table:cifar_asr_allmodel}, low transferability does not necessarily mean that the adversarial example is correctly classified, leaving us the space to further improve the performance of XEnsemble.
These observations are consistent with the result in \cite{papernot2016transferability}, which shows the existence of many weak transferability scenarios across classifiers.

\begin{table*}[ht]
\centering
\scalebox{0.85}{
\small{
\begin{tabular}{|c|c|c|c|c|c|c|c|c|c|c|c|c|}
\hline
\multirow{10}{*}{\rotatebox{90}{ImageNet}} & \multirow{2}{*}{combo} & \multirow{2}{*}{benign} & \multirow{1}{*}{FGSM$_{\infty}$} & \multirow{1}{*}{BIM$_{\infty}$} & \multirow{1}{*}{PGD$_{\infty}$} & \multicolumn{2}{c|}{CW$_\infty$} & \multicolumn{2}{c|}{CW$_2$} & \multicolumn{2}{c|}{CW$_0$}  & \multirow{2}{*}{average} \\ \cline{4-12}
  &  & &  \multicolumn{3}{c|}{UA}  & most & LL & most & LL & most & LL &  \\ \cline{2-13}
 & TM(no defense) & 0.695 & 0.01 & 0 & 0 & 0 & 0.04 & 0 & 0.06 & 0 & 0 & 0.014 \\
%& EnsTrans & 0.715 & 0.41 & 0.6 & 0.8 & 0.91 & 0.82 & 0.92 & 0.76 & 0.86 & 0.76 \\
& VM1 & 0.67 & 0.73 & 0.77 & 0.84 & 0.81 & 0.81 & 0.81 & 0.82 & 0.8 & 0.79 & 0.793 \\
& VM2 & 0.68 & 0.7 & 0.78 & 0.85 & 0.83 & 0.83 & 0.84 & 0.84 & 0.81 & 0.76 & 0.799 \\
& VM3 & 0.67 & 0.78 & 0.84 & 0.83 & 0.85 & 0.83 & 0.83 & 0.84 & 0.83 & 0.8 & 0.825 \\
& VM4 & 0.735 & 0.86 & 0.85 & 0.85 & 0.92 & 0.91 & 0.92 & 0.9 & 0.91 & 0.84 & 0.889 \\
 \cline{2-13}
  & \textbf{XEnsemble-rand} & 0.77 & 0.92 & 0.92 & 0.90 & 0.92 & 0.94 & 0.91 & 0.93 & \textbf{0.92} & 0.95 & 0.923 \\ %\cline{3-15}
  & \textbf{XEnsemble-$\kappa$-rand} & 0.755 & 0.83 & 0.9 & 0.89 & 0.92 & 0.91 & 0.92 & 0.9 & 0.89 & 0.89 & 0.894 \\ %\cline{3-15}
  & \textbf{XEnsemble-best-$\kappa$} & \textbf{0.805} & \textbf{0.94} & \textbf{0.93} & \textbf{0.92} & \textbf{0.96} & \textbf{0.96} & \textbf{0.95} & \textbf{0.95} & 0.91 & \textbf{0.97} & \textbf{0.943} \\ %\hline
  \hline
\end{tabular}
}}
\caption{\small Defense Success Rate measurement of ImageNet under adversarial attacks.}
\label{table:defense_table_imagenet}
\end{table*}

\begin{table*}[ht]
\centering
\scalebox{0.85}{
\small{
\begin{tabular}{|c|c|c|c|c|c|c|c|c|l}
\cline{1-9}
\multicolumn{2}{|c|}{\multirow{3}{*}{ImageNet}} & \multirow{3}{*}{\begin{tabular}[c]{@{}c@{}}test \\ acc\end{tabular}} & \multicolumn{3}{c|}{model ensemble} & \multicolumn{3}{c|}{input-output verification ensemble} &  \\ \cline{4-9}
\multicolumn{2}{|c|}{} &  & XEnsemble-rand & XEnsemble-$\kappa$-rand & XEnsemble-best-$\kappa$ & XEnsemble-rand & XEnsemble-$\kappa$-rand & XEnsemble-best-$\kappa$ &  \\ \cline{4-9}
\multicolumn{2}{|c|}{} &  & VM 1,4 & VM 1,3,4 & VM 1-4 & VM 1,4 & VM 1,3,4 & VM 1-4 &  \\ \cline{1-9}
\multicolumn{2}{|c|}{no attack} & 0.695 & 0.77/0.67/0.10 & 0.755/0.69/0.065 & 0.805/0.67/0.135 & 0.785/0.68/0.105 & 0.745/0.675/0.07 & 0.90/0.715/0.175 &  \\ \cline{1-9}
\multicolumn{2}{|c|}{FGSM} & 0.01 & 0.92/0.77/0.15 & 0.83/0.80/0.03 & 0.94/0.81/0.13 & 0.93/0.74/0.19 & 0.90/0.82/0.08 & 0.96/0.78/0.18 &  \\ \cline{1-9}
\multicolumn{2}{|c|}{BIM} & 0 & 0.92/0.83/0.09 & 0.90/0.85/0.05 & 0.93/0.82/0.11 & 0.90/0.78/0.12 & 0.92/0.85/0.07 & 0.99/0.81/0.18 &  \\ \cline{1-9}
\multicolumn{2}{|c|}{PGD} & 0 & 0.91/0.83/0.08 & 0.89/0.85/0.04 & 0.92/0.82/0.10 & 0.92/0.84/0.08 & 0.91/0.86/0.05 & 0.97/0.82/0.15 &  \\ \cline{1-9}
\multirow{2}{*}{CW$_\infty$} & most & 0 & 0.92/0.86/0.06 & 0.92/0.87/0.05 & 0.96/0.86/0.10 & 0.93/0.79/0.14 & 0.90/0.83/0.07 & 1/0.88/0.12 &  \\ \cline{2-9}
 & LL & 0.04 & 0.94/0.85/0.09 & 0.91/0.85/0.06 & 0.96/0.84/0.12 & 0.94/0.80/0.14 & 0.92/0.87/0.05 & 0.93/0.89/0.04 &  \\ \cline{1-9}
\multirow{2}{*}{CW$_2$} & most & 0 & 0.91/0.85/0.06 & 0.92/0.87/0.05 & 0.95/0.86/0.09 & 0.91/0.79/0.12 & 0.90/0.84/0.06 & 0.96/0.84/0.12 &  \\ \cline{2-9}
 & LL & 0.06 & 0.93/0.88/0.05 & 0.90/0.86/0.04 & 0.95/0.84/0.11 & 0.95/0.79/0.16 & 0.91/0.84/0.07 & 1/0.89/0.11 &  \\ \cline{1-9}
\multirow{2}{*}{CW$_0$} & most & 0 & 0.92/0.84/0.08 & 0.89/0.84/0.05 & 0.91/0.83/0.08 & 0.89/0.79/0.10 & 0.89/0.80/0.09 & 0.99/0.86/0.13 &  \\ \cline{2-9}
 & LL & 0 & 0.95/0.79/0.16 & 0.89/0.83/0.06 & 0.97/0.81/0.16 & 0.94/0.76/0.18 & 0.88/0.77/0.11 & 0.97/0.82/0.15 &  \\ \cline{1-9}
\end{tabular}
}}
\caption{\small Defense details of XEnsemble on ImageNet. Number measured in DSR/PSR/TSR}
\label{table:defense_detail_imagenet}
\end{table*}

{\bf Comparison Study.\/} We further strengthen the robustness performance of XEnsemble with the input augmentation ensemble and output model verification ensembleimaging smoothing
. We compare the effectiveness of input-model verification XEnsemble solutions with 9 existing defense approaches from two categories: defense with no detection and defense with only detection. The former focuses on providing the correct prediction label for adversarial examples and does not integrate the detection function. Therefore, DSR= PSR and TSR=0.
We consider Adversarial Training (AdvTrain~\cite{goodfellow6572explaining}), Ensemble Adversarial Training (EnsTrain~\cite{tramer2017ensemble}), PGD-defense~\cite{madry2017towards}, Defensive Distillation (DefDistill~\cite{papernot2016distillation}), and Input Transformation Ensemble (EnsTrans~\cite{xie2017mitigating}) in the defense with no detection category. defense with only detection aims at flagging whatever input that is identified as adversarial example and cannot provide a repaired label for these flagged input, i.e. DSR = TSR and PSR=0. We consider Feature Squeezing (FS~\cite{xu2017feature}), MagNet~\cite{meng2017magnet}, NIC~\cite{ma2019nic} and LID~\cite{ma2018dawn}.

For adversarial training, we use the adversarial training examples generated from FGSM with random $\theta$ within $[0, 0.0156]$. In adversarial training ensemble, the adversarial example is generated from 3 of the 6 models(TM+ 5VMs).  For defensive distillation, the temperature is set to 50. The input transformation ensemble computes multiple randomly cropped-and-padded input image. The ensemble size is set to 10 while the crop size is 28. The comparison results are provided in Table~\ref{table:comparison_table}. We make two interesting observations: (1) On both adversarial attacks and OOD input, the XEnsemble defense approach consistently and significantly outperforms those prevention-only defense methods regardless whether it is the random ensemble from the base model pool, or the random $\kappa$ ensemble from the pool of $\kappa$ ranked ensemble list, or the best $\kappa$ ensemble. (2) Although defense with only detection methods can have high defense success rate, they are passive defenses by resigning from guaranteeing robustness, incapable of surviving the ML component for its routine function under attack. While defense with only detection methods  can be useful in sensitive applications like malware detection, such defenses are not suitable for mission-critical systems that cannot tolerate real-time interruptions. Our method can achieve comparable defense success rate and perform extremely well on the OOD input. (3)
the best $\kappa$ ensemble is effective across all 11 attacks and the winner for all 8 targeted attacks. Another merit of the input-model verification XEnsemble is that many of the augmentation techniques are non-differentiable. Therefore, the input ensemble cannot be jointly optimized with multiple models to perform the cross-layer attack.

%Although some work~\cite{athalye2018obfuscated} points out that the differentiable Expectation-over-Transformation could mimic the behavior of input augmentation and could be used to simulate the input layer, further randomness can be injected to either dynamically changing the parameters or the type of the input augmentation. It would be difficult to design a dynamic Expectation-over-Transformation simulation that can simultaneously decode the augmentation techniques in use.

\begin{table*}[ht]
\centering
\scalebox{0.80}{
\small{
\begin{tabular}{|c|c|c|c|c|c|c|c|c|c|c|c|c|c|}
\hline
\multirow{3}{*}{method} & \multirow{3}{*}{setting} & \multicolumn{6}{c|}{CIFAR-10} & \multicolumn{6}{c|}{CIFAR-100} \\ \cline{3-14}
 & & \multicolumn{3}{c|}{TinyImageNet} & \multicolumn{3}{c|}{LSUN} & \multicolumn{3}{c|}{TinyImageNet} & \multicolumn{3}{c|}{LSUN} \\ \cline{3-14}
 &  & DSR & DError & AUROC & DSR & DError & AUROC & DSR & DError & AUROC & DSR & DError & AUROC \\ \hline
baseline & \cite{hendrycks2016baseline} & 0.82 & 0.225 & 0.941 & 0.854 & 0.098 & 0.954 & 0.27 & 0.39 & 0.71 & 0.258 & 0.396 & 0.707\\ \hline
\multirow{3}{*}{\begin{tabular}[c]{@{}c@{}}ODIN \\\cite{liang2017enhancing} \end{tabular}} & T=1000, N=0.0014 &0.938 & 0.056 & 0.985 & 0.962 & 0.055 & 0.992 & 0.574 & 0.238 & 0.88 & 0.576 & 0.237 & 0.885 \\
 & T=1, N=0.0007 & 0.876 & 0.087 & 0.969 & 0.906 & 0.073 & 0.973 & 0.546 & 0.252 & 0.863 & 0.548 & 0.251 & 0.864 \\
 & T=1000, N=0.01 & 0.542 & 0.254 & 0.884 & 0.716 & 0.267 & 0.926 & 0.246 & 0.402 & 0.705 & 0.22 & 0.415 & 0.694 \\ \hline
Mahalanobis & \cite{lee2018simple} & 0.95 & 0.05 & 0.988 & 0.976 & 0.047 & 0.993 & 0.894 & 0.078 & 0.974 & 0.928 & 0.061 & 0.98\\ \hline
Leave-out & \cite{vyas2018out} & 0.974 & 0.038 & 0.993 & 0.996 & 0.037 & 0.998 & 0.85 & 0.1 & 0.963 & 0.874 & 0.088 & 0.97 \\ \hline

FS & \multirow{4}{*}{\begin{tabular}[c]{@{}c@{}}Defense \\ with only \\ detection\end{tabular}} & 0.8016 & 0.1192 & 0.9226 & 0.7735 & 0.1358 & 0.9012 & 0.8534 & 0.0948 & 0.9693 & 0.8435 & 0.0962 & 0.9573 \\
MagNet &  & 0.9276 & 0.0547 & 0.9776 & 0.9678 & 0.0326 & 0.9835 & 0.4535 & 0.2952 & 0.8549 & 0.4217 & 0.3067 & 0.8326 \\
NIC &  & 0.9053 & 0.0694 & 0.9742 & 0.9212 & 0.0579 & 0.9726 & 0.8012 & 0.1174 & 0.9331 & 0.7992 & 0.1147 & 0.958 \\
LID &  & 0.8772 & 0.0774 & 0.9683 & 0.9487 & 0.0402 & 0.9792 & 0.7688 & 0.1331 & 0.9385 & 0.7876 & 0.1281 & 0.9484\\ \hline

XEnsemble-rand &   \multirow{3}{*}{our method} & 0.996 & 0.002 & 0.999 & 0.986 & 0.007 & 0.999 & 0.998 & 0.001 & \textbf{$\approx$ 1} & 0.99 & 0.005 &  \textbf{$\approx$ 1} \\
XEnsemble-$\kappa$ &  & \textbf{0.998} & \textbf{0.001} & \textbf{$\approx$ 1} & 0.992 & 0.004 & $\approx$ 1 & 0.998 & 0.001 & \textbf{$\approx$ 1} & 0.992 & 0.004 & \textbf{$\approx$ 1} \\
XEnsemble-best &  & \textbf{0.998} & \textbf{0.001} & \textbf{$\approx$ 1} & \textbf{0.996} & \textbf{0.002} & \textbf{$\approx$ 1} & \textbf{$\approx$ 1} & \textbf{$\approx$ 0} & \textbf{$\approx$ 1} &\textbf{0.996} & \textbf{0.002} & \textbf{$\approx$ 1}\\ \hline
\end{tabular}
}}
\caption{\small  DSR, Detection error and AUROC Comparison of XEnsemble with representative out-of-distribution data detection techniques. For methods that are not XEnsemble, the DSR is measured by (1-\textbf{FPR at 0.95 TPR}).}
\label{table:ood_compare}
\end{table*}

\subsection{Defending Adversarial Attacks on ImageNet}

We next evaluate the effectiveness of XEnsemble on ImageNet and provide the results in Table~\ref{table:defense_table_imagenet}. For ImageNet models,
the XEnsemble-best-$\kappa$ ensemble consists of the target model the VM 1,2,3,4 and provides the best defense for 8 out of 9 attacks.
Although the single-model accuracy of ImageNet is not as high as CIFAR-10, the discrepancy between models plays an important role in the proposed model verification ensemble defense when auto-verifying the input. Particularly as shown in Table~\ref{table:defense_detail_imagenet}, the defense success rate on ImageNet not only comes from the competitive prediction accuracy(around 70\%) of the individual model but also results from the highly diverse predictions of the same input under different models. Ensemble of these verification models brings a prediction accuracy over 80\% on the selected 100 attack inputs for each attack while flagged inputs contribute to an additional 10\% TSR to the DSR. Meanwhile, Table~\ref{table:defense_detail_imagenet} also shows the robustness improvement brought by the input defense layer. Note that the JSMA attack is not included for ImageNet due to its overwhelming overhead on the heuristic search of the pixel space in high-resolution images during the attack generation. We also observe that the benign test accuracy of all three XEnsemble ensembles and the input transformation ensemble defense are higher than that of the target model, showing the power of ensemble learning over individual members of the ensemble. Another merit of the proposed defense is that it does not require a modification nor re-training of the target DNN model under protection, which can be very expensive for large-scale images such as ImageNet.

The experiments also indicate that the ensemble defense success rate does not have a high correlation with the size of the ensemble, thus high diversity ensemble of one size may have a high accuracy comparable with those ensembles of other ensemble sizes. Although the effects and benefits of diversity
ensemble is witnessed in our experiment and many other researches, there are many issues worth further investigation, including the quantification of ensemble diversity with theoretical formulation and the trade-off between diversity and accuracy in ensemble generation.

%and Bagging Ensemble (BagEns~\cite{strauss2017ensemble})

\subsection{Effectiveness of Detecting OOD Inputs}

We perform the out-of-distribution input detection experiment on both CIFAR-10 and CIFAR-100. We use the DenseNet CIFAR-100 model with  76.65\% benign accuracy as the target model and create the ensemble pool for verification ensembles in a similar manner as those in the previous set of experiments on CIFAR-10 and ImageNet. The CIFAR-100 model pool consists of a number of Wide-ResNet models: WRN-28-10 with 80.75\% benign accuracy, WRN-28-10-dropout with 81.15\% accuracy, WRN-40-4 with 79.27\% accuracy, and WRN-40-10 with 81.7\% accuracy. The first parameter is the model depth and the second parameter is the widen factor.
For both CIFAR-10 and CIFAR-100, We conduct the OOD detection experiment by feeding the entire TinyImageNet data and LSUN data, each 10,000 images, to a DenseNet classifier trained for CIFAR-10 and CIFAR-100.  For both TinyImageNet and LSUN, we down-sample the data to size 32*32 to fit the prediction classifier. Similar to the adversarial example, we set the ranking confidence level to 0.5, meaning that we will flag and reject the input as long as more than half of the model cannot agree on their prediction.

%~\cite{zagoruyko2016wide}

We evaluate the effectiveness of XEnsemble against out-of-distribution inputs by comparing it with four existing representative approaches:
(1) the baseline OOD detection~\cite{hendrycks2016baseline},
(2) the ODIN for OOD image detection~\cite{liang2017enhancing},
(3) the Mahalanobis-distance based OOD detection~\cite{lee2018simple} and
(4) the Leave-out classifier ensemble based OOD detection~\cite{vyas2018out}. To illustrate the influence of hyperparameters in ODIN, we include the measurement of using three different choices of ODIN parameters. Note that XEnsemble defense is independent of both adversarial attacks and out-of-distribution examples. Thus, unlike existing OOD defenses that rely on setting certain detection thresholds based on the out-of-distribution examples, XEnsemble verification ensemble approach can achieve a high detection success rate independent of any OOD-data specific threshold, or the noise level and temperature scaling in ODIN. Thus, unlike  the baseline OOD detector, ODIN, and Mahalanobis-distance based detector, there is no \textbf{FPR at 0.95 TPR} measurement for Xensemble. \textbf{FPR at 0.95 TPR} is the probability that a negative (out-of-distribution) example is misclassified as positive (in-distribution) when the true positive rate (TPR) is higher than 95\%. We compare the DSR of our method in detecting OOD input in Table~\ref{table:ood_compare} and compare it with the detection accuracy of other methods when their TPR is 0.95.

Given the target model, since there is no correct prediction for the out-of-distribution examples, the goal of our model verification ensemble defense is to achieve a high defense success rate (DSR) by maximizing the detection success rate (TSR). Table~\ref{table:ood_compare} shows the results. The XEnsemble model verification ensemble defense is able to effectively improve the detection performance for out-of-distribution examples and has superior defense performance against OOD input in all three evaluation metrics. Specifically, the detection success rate for the XEnsemble defense on CIFAR-100 is better than that on CIFAR-10. This is partly attributed to the richer choice of prediction classes in CIFAR-100 such that the wrongly predicted class for each OOD example is scattered more in the prediction label space for CIFAR-100 classifiers.

\section{Limitation and Improvement}

% We have demonstrated that the XEnsemble is effective against adversarial examples and out-of-distribution inputs under the black-box defense threat model in which the adversary has no knowledge of the composition of the defense system (recall Section~\ref{section3.1}).
% However, as reported in the literature, existing defenses fail to generalize over advanced threat models~\cite{goodfellow2018defense}, and they are completely broken under the white box threat model in which the attacker knows everything about the target system (both the target model TM, and the composition of the defense)~\cite{he2017adversarial}. Thus,
% the proposed XEnsemble defense has its limitation: it may be broken by the cross-model ensemble attacks when a partial knowledge (grey-box) or the full knowledge (white-box) of the defense system is exposed to the adversary, such as insider snooping. For example, synthesized robust adversarial example \cite{athalye2017synthesizing} and ensemble targeted adversarial example \cite{liu2016delving} take gradient information from multiple models to perform the cross-model ensemble attack~\cite{li2018learning,li2019regional,zhou2018transferable,xie2018improving}, aiming to create transferable adversarial perturbations across models.

We have demonstrated that the XEnsemble is effective against adversarial examples and out-of-distribution inputs under the black-box defense threat model in which the adversary has no knowledge of the composition of the defense system (recall Section~\ref{section3.1}).
In this section, we evaluate the insider attack scenarios, in which the adversary has partial or full knowledge of the XEnsemble system. We refer to these two threat scenarios as grey box threats and white box threats as defined in Section~\ref{section3.1}. As reported in~\cite{goodfellow2018defense,he2017adversarial}, existing defenses fail to generalize over black box threats in the sense that some defense methods work well for certain set of attacks but fail under other types of attacks. Moreover, most existing defenses are broken under the white box threat model in which the attacker has full knowledge of the target system (both the target victim model (TM), and the composition of the defense ensemble). In this section, we evaluate the limitation of
the proposed XEnsemble approach: When and how it may be broken by the cross-model ensemble attacks when a partial knowledge (grey-box) or the full knowledge (white-box) of the defense system is exposed to the adversary, such as insider snooping.

\begin{table}[t]
\centering
\scalebox{0.72}{
\small{
\begin{tabular}{|c|c|c|c|c|c|c|c|c|}
\hline
\multirow{2}{*}{\begin{tabular}[c]{@{}c@{}}ensemble \\ attack\end{tabular}} & \multicolumn{2}{c|}{CW$_\infty$} & \multicolumn{2}{c|}{CW$_2$} & \multicolumn{2}{c|}{CW$_0$} & \multirow{2}{*}{\begin{tabular}[c]{@{}c@{}}avg \\ DSR\end{tabular}} & \multirow{2}{*}{\begin{tabular}[c]{@{}c@{}}exposed \\ model\end{tabular}} \\ \cline{2-7}
 & most & LL & most & LL & most & LL & &  \\ \hline
TM & 0 & 0 & 0 & 0 & 0 & 0 & 0 & TM \\ \hline
DM 1 & 0 & 0 & 0 & 0 & 0 & 0 & 0 & DM 1 \\ \hline
DM 2 & 0 & 0 & 0 & 0 & 0 & 0 & 0 & DM 2 \\ \hline
DM 4 & 0 & 0 & 0 & 0 & 0 & 0 & 0 & DM 4 \\ \hline
DM 5 & 0 & 0 & 0 & 0 & 0 & 0 & 0 & DM 5 \\ \hline
black-box & 1 & 0.98 & 0.99 & 0.97 & 0.92 & 0.91 & 0.962 & TM \\ \hline
grey-rand & 0.77 & 0.85 & 0.88 & 0.94 & 0.74 & 0.87 & 0.842 & TM, DM 1,2 \\ \hline
grey-fix & 0.49 & 0.58 & 0.66 & 0.78 & 0.47 & 0.65 & 0.605 & TM, DM 1,2 \\ \hline
white-rand & 0 & 0 & 0 & 0 & 0 & 0 & 0 & \multirow{5}{*}{TM, DM 1-7}\\ \cline{1-8}
white-fix & 0 & 0 & 0 & 0 & 0 & 0 & 0 &   \\ \cline{1-8}
DistPerturb & 0.922 & 1.189 & 0.725 & 0.972 & 1.614 & 2.046 & - &  \\ \cline{1-8}
DistPercept & 21.76 & 33.78 & 9.81 & 18.12 & 12.58 & 23.77 & - &  \\ \cline{1-8}
Time(s) & 350.4 & 362.5 & 12.43 & 17.52 & 402.8 & 445.3 & - &  \\ \hline
\end{tabular}
}}
\caption{\small Defense success rate under model verification ensemble CW attack on CIFAR-10.}
\label{table:online}
%\vspace{-0.4cm}
\end{table}

{\bf Experiment Setup.\/} For grey-box attacks, we consider two kinds of the adversaries: (1) the attacker knows some of the models in the entire ensemble team of models and the ensemble is fixed ({\bf grey-fix}); and (2) the attacker knows some of the models in the entire base model pool and the ensemble is randomly selected for every query ({\bf grey-rand}). Similarly, we consider two kinds of white-box adversaries: (3) the attacker knows all of the base models in the defense system and the ensemble team used for prediction is fixed ({\bf white-fix}); and (4) the attacker knows all of the baseline models and the ensemble team is randomly selected for every query ({\bf white-rand}).
We conduct the ensemble CW attack on CIFAR-10 using the target DenseNet model and the 5 verification models listed in Table 3. For the grey box scenarios, we assume that the adversary knows 2 out of the 5 verification models (e.g., VM 1, 2). For the white-box scenarios, we assume that all 5 VMs are known to the adversary. In both cases, ensemble models with a ensemble size of 4 will be chosen. Table~\ref{table:online} shows the results, measured by the defense success rate DSR on each verification model or ensemble option. It also includes the attack costs for the white-box scenario in terms of perturbation distance, perception distance and time cost.

{\bf Empirical Analysis.} (1) The robustness of our model verification defense approach deteriorates as the adversary increases his knowledge of the composition of the defense system. XEnsemble defense achieves the robustness in terms of average DSR of 84.2\% for grey-rand scenario, compared to 60.5\% for grey-fix. (2) Our approach fails completely with 0\% DSR for white-box scenarios (both white-fix and white-rand). Table~\ref{table:online} also lists the attack costs for the white-box scenario. We see that the cross-model ensemble attack shows the increased perturbation distance, increased perception distance and significantly increased time cost when compared to Table~\ref{table:attacks}. We are encouraged by this preliminary study result for two reasons. First, both grey-box and white-box attacks are only experimented with our output model verification ensemble layer with a small pool of 5 base models on CIFAR-10. We did not incorporate the input-denoising layer ensemble, which could add $k$ times more alternative verifiers and also make the ensemble attack harder to generalize. Second, the output model verification ensemble defense used in this set of experiments did not leverage Kappa diversity, which is another defense parameter, if unknown to the grey-box adversary, would boost the robustness of our defense approach.

{\bf Takeaway.} we conjecture that (1) extending the size of the baseline candidate set for output-layer verification model ensemble is an important optimization since it can significantly increase the cost for adversaries under grey-box or white-box scenarios; (2) by incrementally adding new models of high test accuracy to expand the baseline candidate model pool over time, we can substantially boost the robustness of XEnsemble defense against grey-box and white-box attacks, turning the white-box attack into grey-box; and (3) by introducing randomization to the input denoising layer, we will have the opportunity to further strengthen the robustness and survivability of our defense approach.

\section{Conclusion}
We have presented XEnsemble, an input verification and output verification ensemble defense methodology with three novel features. First, XEnsemble is capable of defending a DNN model under protection by utilizing both model-structure diversity and model-disagreement diversity.
Second, XEnsemble improves the robustness of the target DNN model by providing auto-repairing and auto-detection capability in the presence of adversarial examples and out-of-distribution examples.  Finally, the proposed XEnsemble defense is attack independent. It does not require re-training the target DNN model and can generalize over attack algorithms as well as different out-of-distribution datasets. Evaluated over eleven attack algorithms and two out-of-distribution datasets, we show that XEnsemble can achieve a high defense success rate on both adversarial and OOD inputs.

Our research continues along two directions: (1) We plan to introduce randomization in both input denoising ensemble layer and output model verification layer to add additional robustness to the XEnsemble framework against insider attacks under grey box and white box threat models to defense systems. (2) We plan to extend the XEnsemble approach to other modalities, such as video, audio, and text against adversarial attacks, such as~\cite{wei2020framework}.

\vspace{0.6cm}

% \section*{Acknowledgement}
%This work is partially sponsored by NSF CISE SaTC grant 1564097 and an IBM faculty award.
\textbf{Acknowledgement.}
The authors acknowledges a partial support by the National Science Foundation under Grants NSF 2038029, NSF 1564097, and an IBM faculty award.

\bibliographystyle{IEEEtran}
\bibliography{main}

\begin{IEEEbiography}
[{\includegraphics[width=1in,height=1.25in,clip,keepaspectratio]{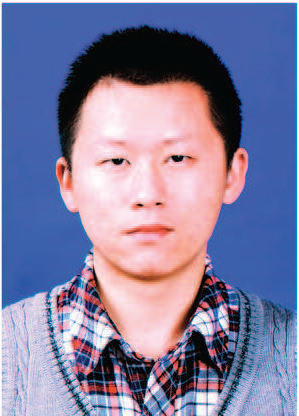}}]{Wenqi Wei}
is currently pursuing his Ph.D. in the School of Computer Science, Georgia Institute of Technology, where he is advised by Prof.~Ling Liu. He received his B.E. degree from the School of Electronic Information
and Communications, Huazhong University of Science
and Technology. His research interests include data privacy, security, machine learning, and big data analytics.
\end{IEEEbiography}

\vspace{-20pt}

\begin{IEEEbiography}
[{\includegraphics[width=1in,height=1.25in,clip,keepaspectratio]{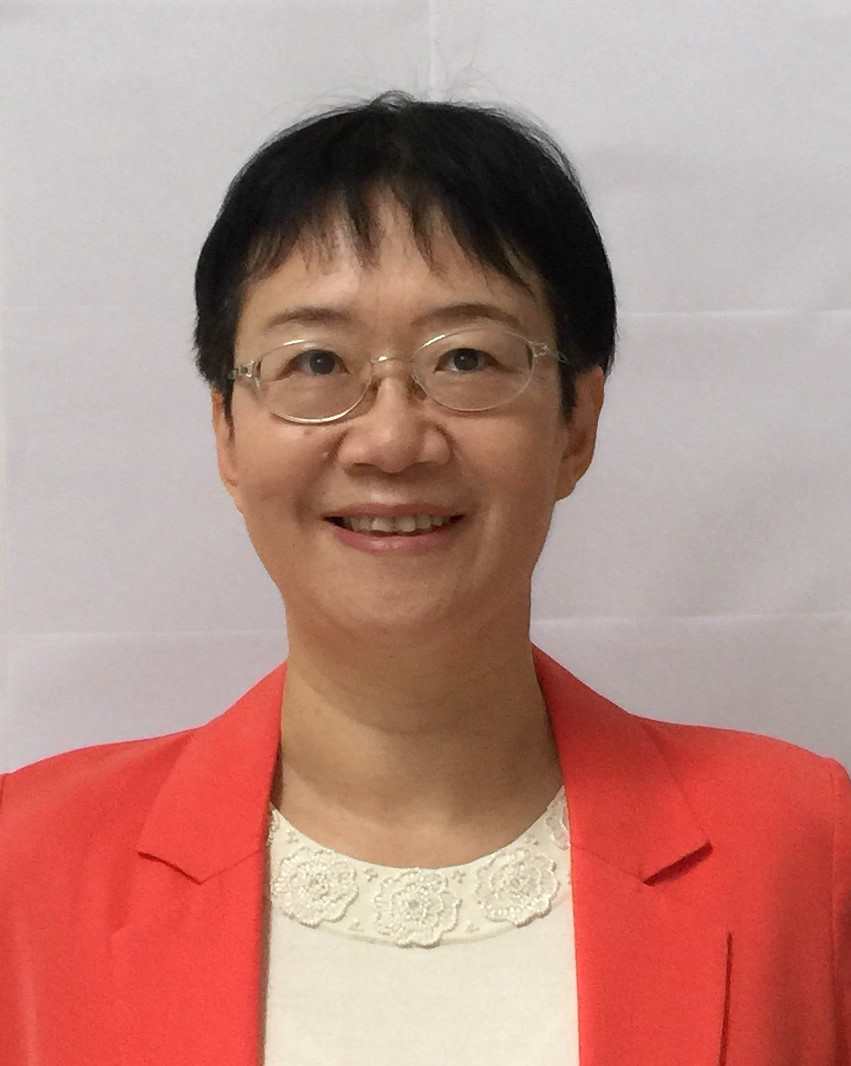}}]{Ling Liu}
is a professor in the School of Computer Science, Georgia Institute of Technology. She directs the research programs in the Distributed Data
Intensive Systems Lab (DiSL). She is an elected IEEE fellow, a recipient of the IEEE Computer Society Technical Achievement Award in 2012, and a recipient of the best paper award from a dozen of top venues, including ICDCS 2003, WWW 2004, 2005 Pat Goldberg Memorial Best Paper Award, IEEE Cloud 2012, IEEE ICWS 2013, ACM/IEEE CCGrid 2015, and IEEE Symposium on BigData 2016. In addition to serving as the general chair and PC chair of numerous IEEE and ACM conferences in data engineering, she has served on the editorial board of over a dozen international journals. Her current research is primarily sponsored by NSF, IBM, and Intel.
\end{IEEEbiography}

\end{document}